\documentclass{article}

\usepackage{arxiv}

\usepackage[utf8]{inputenc} 
\usepackage[T1]{fontenc}    
\usepackage{hyperref}       
\usepackage{url}            
\usepackage{booktabs}       
\usepackage{amsfonts}       
\usepackage{nicefrac}       
\usepackage{microtype}      

\usepackage{lipsum}         
\usepackage{graphicx}
\usepackage{natbib}
\usepackage{doi}
\usepackage{cite}
\usepackage{subfig}
\usepackage{textcomp}
\usepackage{xcolor}
\def\BibTeX{{\rm B\kern-.05em{\sc i\kern-.025em b}\kern-.08em
    T\kern-.1667em\lower.7ex\hbox{E}\kern-.125emX}}
\usepackage{amsmath,graphicx,amssymb,amsfonts}
\usepackage{cleveref}       

\usepackage{booktabs}
\usepackage{multirow}

\usepackage{algorithm}
\usepackage[noend]{algpseudocode}
\makeatletter
\def\BState{\State\hskip-\ALG@thistlm}
\makeatother

\title{Graph Federated Learning for CIoT Devices in Smart Home Applications}

\author{ \href{https://orcid.org/0000-0001-6489-5235}{\includegraphics[scale=0.06]{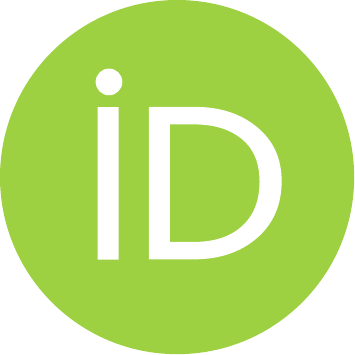}\hspace{1mm}Arash Rasti-Meymandi} \\
	Department of Electrical and Computer Engineering\\
	University of Toronto\\
	Toronto, Canada \\
	\texttt{arash.rasti@mail.utoronto.ca} \\
	\And
	\href{https://orcid.org/0000-0001-9380-3248}{\includegraphics[scale=0.06]{orcid.pdf}\hspace{1mm}Seyed Mohammad Sheikholeslami} \\
	Department of Electrical and Computer Engineering\\
	University of Toronto\\
	Toronto, Canada \\
	\texttt{sm.sheikholeslami@mail.utoronto.ca} \\
	\And
	\href{https://orcid.org/0000-0002-2608-6100}{\includegraphics[scale=0.06]{orcid.pdf}\hspace{1mm}Jamshid Abouei} \\
	Department of Electrical Engineering\\
	Yazd University\\
	Yazd, Iran\\
	\texttt{abouei@yazd.ac.ir} \\
	\And
	\href{https://orcid.org/0000-0003-3647-5473}{\includegraphics[scale=0.06]{orcid.pdf}\hspace{1mm}Konstantinos N. Plataniotis} \\
	Department of Electrical and Computer Engineering\\
	University of Toronto\\
	Toronto, Canada \\
	\texttt{kostas@comm.utoronto.ca} \\
}

\renewcommand{\shorttitle}{Graph Federated Learning for CIoT Devices in Smart Home Applications}

\hypersetup{
pdftitle={Rasti_GFIoTL},
pdfsubject={q-bio.NC, q-bio.QM},
pdfauthor={David S.~Hippocampus, Elias D.~Striatum},
pdfkeywords={First keyword, Second keyword, More},
}

\begin{document}
\maketitle

\begin{abstract}
	This paper deals with the problem of statistical and system heterogeneity in a cross-silo Federated Learning (FL) framework where there exist a limited number of Consumer Internet of Things (CIoT) devices in a smart building. \textcolor{black}{We propose a novel Graph Signal Processing (GSP)-inspired aggregation rule based on graph filtering dubbed ``G-Fedfilt''\footnote{The implementation is available at https://github.com/FL-HAR/Graph-Federated-Learning-for-CIoT-Devices.git}. The proposed aggregator enables a structured flow of information based on the graph's topology. This behavior allows capturing the interconnection of CIoT devices and training domain-specific models.} The embedded graph filter is equipped with a tunable parameter which enables a continuous trade-off between domain-agnostic and domain-specific FL. In the case of domain-agnostic, it forces G-Fedfilt to act similar to the conventional Federated Averaging (FedAvg) aggregation rule. The proposed G-Fedfilt also enables an intrinsic smooth clustering based on the graph connectivity without explicitly specified which further boosts the personalization of the models in the framework. \textcolor{black}{In addition, the proposed scheme enjoys a communication-efficient time-scheduling to alleviate the system heterogeneity. This is accomplished by adaptively adjusting the amount of training data samples and sparsity of the models' gradients to reduce communication desynchronization and latency.} Simulation results show that the proposed G-Fedfilt achieves up to $3.99\% $ better classification accuracy than the conventional FedAvg when concerning model personalization on the statistically heterogeneous local datasets, while it is capable of yielding up to $2.41\%$ higher accuracy than FedAvg in the case of testing the generalization of the models. Furthermore, the proposed communication optimization scheme can boost the framework's efficiency by reducing \textcolor{black}{the computation, communication desynchronization, and latency up to $70.21\%$, $99.65\%$, and $44.61\%$, respectively,} at the cost of $0.36\%$ accuracy and under the system heterogeneity\footnote{This article has been accepted for publication in IEEE Internet of Things Journal. This is the author's version which has not been fully edited and
content may change prior to final publication. Citation information: DOI: 10.1109/JIOT.2022.3228727}.
\end{abstract}


\section{Introduction} \label{Se:Intro}
Within the past few years, there has been tremendous growth in Consumer Internet of Things (CIoT) and personal smart devices with powerful communication and computation capabilities \citep{poushter2016smartphone}. These devices often share a common goal when it comes to applications such as Human Activity Recognition (HAR), object detection and object recognition. In this regard, they have Machine Learning (ML) models embedded inside to make decisions based on the training data and without being explicitly programmed to act. For the ML model to perform well in real-world scenarios, training data should be supplied by the clients themselves. However, these data are often restricted by end-users due to privacy concerns. In addition, as the number of edge devices increases, so does the uplink communication of the data which in return, causes a large latency and a communication overhead due to the bandwidth limitations~\citep{chiang2016fog}. Recently, Federated Learning (FL), an advanced distributed ML algorithm, has been recognized as a promising solution for the above challenges \citep{mcmahan2017communication,gafni2021federated}. FL has gained an exponential attraction in both wireless communication and signal processing communities. The key idea behind FL is to distribute the training computation of an ML model at the edge and hence, preserve the privacy of end users' data while conveying less information to the server in each round of communication \citep{li2020federated}. In conventional FL, there are several iterations for model convergence. At each iteration, end users produce local models on their available local datasets. The models are then transmitted to a server for aggregation. Finally, the aggregated model is shared among all devices. This process is repeated until model convergence. Although FL can perform the joint training in a distributed manner, \textcolor{black}{there are still some actively researched challenges that are needed to be addressed before their employment in smart home applications. Some of these challenges include:}

\textcolor{black}{\textbf{Statistical Heterogeneity:} The ubiquity of CIoT devices has brought about a plethora of high-performance machines in smart homes, including smartphones, smartwatches, and smart speakers. Such devices accumulate various information from the clients; however, the collected data is often statistically biased to whom the device belongs. For instance, one client performs a specific gesture by moving his/her hand in a particular manner; while another client makes the same gesture, however, differently and less frequently. As a result, the data samples collected at each smart home differ from each other in two aspects; $i)$ they are acquired from different clients and edge devices resulting in a feature distribution skew, and $ii)$ the labels of data samples are varied from one client or device to another, thus creating a label distribution skew \citep{kairouz2019advances}. These biased characteristics among the local datasets are regarded as ``\textit{statistical heterogeneity}''. The effect is a decrease in the convergence time and the accuracy of the FL model. On the other hand, global aggregations in most conventional FL cannot take statistical heterogeneity into account. Therefore, they are not suitable for FL in smart home applications. Note that conventional aggregation rules in FL such as Federated Averaging (FedAvg) or some of its variants \citep{reddi2020adaptive} are considered global aggregations. They primarily capture the general features rather than personal structures in data; in other words, they are agnostic to domain specifics.}

\textcolor{black}{\textbf{System Heterogeneity:} Edge devices over the CIoT network have diverse computational powers and network connectivity. For example, most smartphone brands nowadays are equipped with high-speed CPU chipsets much more computationally potent than smartwatches. Consequently, the training time of local models on these two devices will significantly vary. Besides, even devices with similar capabilities and brands may show performance fluctuation over time. Such diversity in computation and communication performance among devices is known as ``\textit{system heterogeneity''}. This problem intensifies the asynchronicity of the training during the aggregation which in turn, imposes a large latency and desynchronization in each round of communication. In practice, more valuable power resources and time are consumed.}

\begin{figure*}[t]
\centerline{\includegraphics[scale=.3]{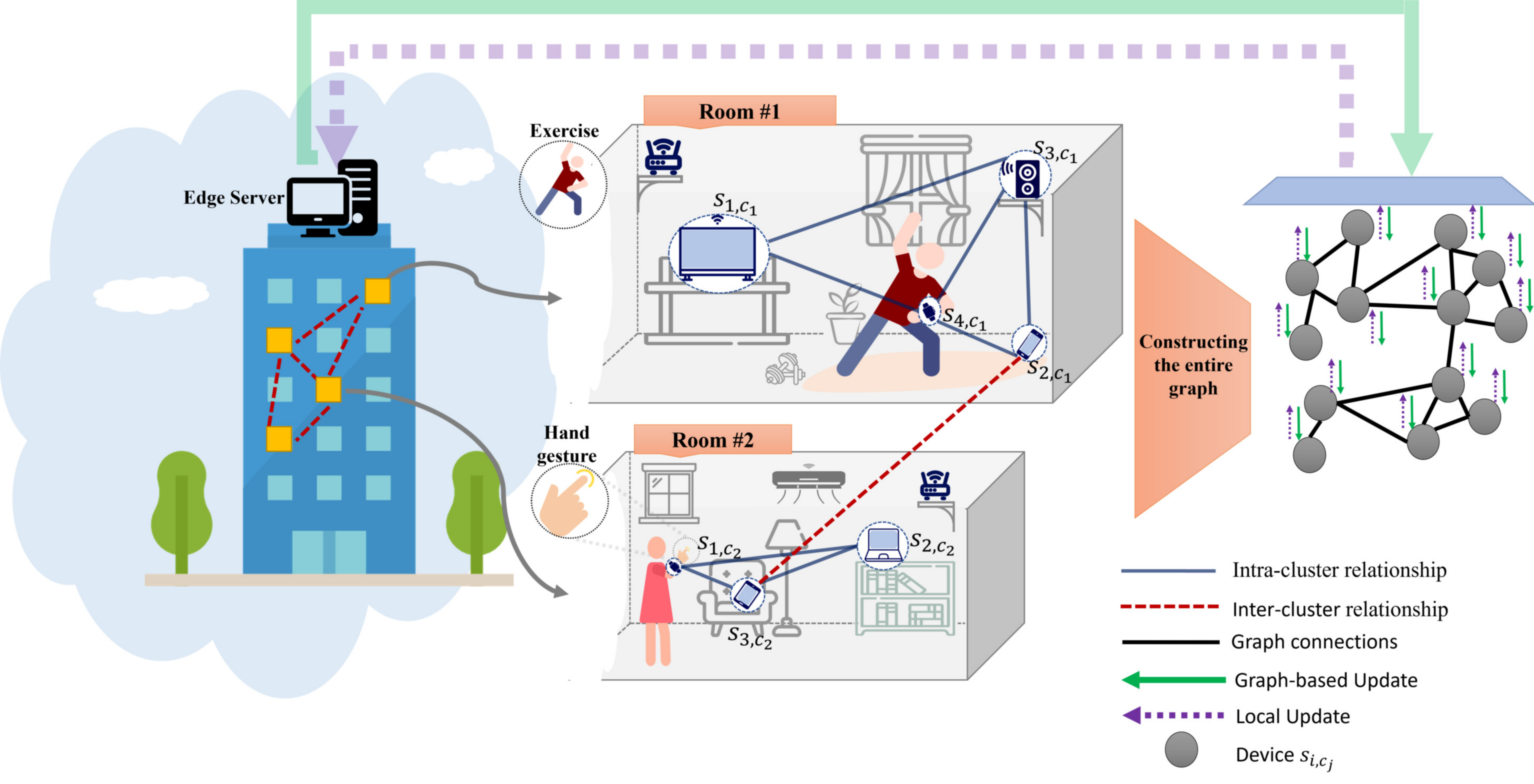}}
\caption{\textcolor{black}{An overview of smart homes with multiple smart devices: Each device has an ML model for a common application. There are intra/inter-cluster relationships among devices (blue and red dashed lines) which are captured by a graph. The models will then be trained collaboratively based on the graph.} } 
\label{fig:HidFL}
\end{figure*}

 \textcolor{black}{There is a surge in describing the intrinsic connections of interactive systems by traditional mathematical representations such as graphs. However, the data supported by such graphs require a new processing mechanism beyond the classical Signal Processing (SP) algorithms. In other words, a unified approach is needed to analyze and extract useful information from the irregularities yet meaningful connections on graphs.} In this regard, Graph Signal Processing (GSP), a generalization of classical SP, aims to handle the graph-structured data and open a new path to better data processing~\citep{dong2020graph}. In addition, it is rather intuitive to presume an underlying relationship between smart devices in an FL framework. This interconnection stems from both the data and the physical hardware of devices and it can be represented by a graph. Thus, GSP can be employed to further benefit from the inherited relationship among smart CIoT devices and better train ML models in a distributed learning paradigm such as FL. \textcolor{black}{As a motivating example, consider the application of HAR or gesture recognition in a smart building with multiple smart homes illustrated in Fig. \ref{fig:HidFL}. Some large-scale activities, such as walking, might be recognizable via a smartphone. However, there are many other large or small-scale activities and gestures demanding an ensembled decision collected from multiple smart devices. Let each device be equipped with an ML model. The inference they make is highly dependent on how they are trained. It is not a far-fetched idea to assume an intrinsic relationship among these devices as they collect data from a specific client (the intra-connections in Fig. \ref{fig:HidFL}). Furthermore, this client might also have some relationship with another client in the building, thus creating inter-connections between devices. To account for such connectivity, one can use a graph. The connectivity could be based on data correlation, hardware specification similarities, or geometrical distances. In such a scenario, training the ML models will be dependent based on the graph. Therefore, the exchange of gradients of the models in the aggregation phase will be handled not in an agnostic but in a more structured fashion. The result is, ML models learn more relatively in the training process and consequently, make a decision more accurately in the inference stage.}
\subsection{Literature Review}
Since its inception in \citep{mcmahan2017communication}, there has been lots of research on addressing the statistical heterogeneity in FL due to the Non-identically independent distributed (Non-i.i.d.) and unbalanced datasets among devices~\citep{khaled2020tighter,abad2020hierarchical,li2019fair}. However, the distribution skewness among edge devices might prevent the global model to converge to an optimum point \citep{li2019convergence}. To tackle this problem, several works have been proposed to find a more personalized model for each device, including $i)$ fine-tuning, where the idea is to fine-tune the model based on the local dataset \citep{deng2020adaptive,mansour2020three,liang2020think},  $ii)$ federated transfer learning, where they freeze the globally trained models and personalize deeper layers of the deep models by retraining them on the local dataset~\citep{chen2020fedhealth,feng2020pmf,arivazhagan2019federated}, $iii)$ federated meta-learning, where a model trained to handle many tasks (e.g., edge devices) is to be trained on a small local dataset with few gradient iterations to get an essence of personalization~\citep{jiang2019improving,fallah2020personalized,khodak2019adaptive}, and $iv)$ federated multi-task learning, where each edge device is trained collaboratively based on others' information and its own update~\citep{corinzia2019variational,smith2017federated}.

 Although all the above researches have conducted a level of personalization for edge devices, they merely considered them as independent devices. In other words, edge device relationship has been underrated within their algorithms. Since its introduction, multiple GSP-inspired algorithms have been developed to handle graph-structured data in different areas such as spectral clustering \citep{tremblay2020approximating} and action recognition \citep{hao2021hypergraph}. There have also been some recent studies on graphs in FL known as ``Federated Graph Learning''~\citep{he2021fedgraphnn}. Most of the current relevant research including \citep{sajadmanesh2021locally,jiang2020federated,wu2021fedgnn,meng2021cross} aims to train a Graph Neural Network (GNN), an advanced GSP-inspired ML model, using FL. Authors in \citep{mei2019sgnn,zheng2021asfgnn} used the graph as an auxiliary representation for improving node classification. In \citep{wang2020graphfl}, authors used FL to train a semi-supervised graph-based node classification in a real-world application. To the best of our knowledge, the only work in the community that utilizes graphs as a way of mitigating statistical heterogeneity is the work in \citep{caldarola2021cluster}. They split the model's weights into two parts of global and local layers where the latter one is trained on shared information provided by a Graph Convolutional Network (GCN). 

In the case of system heterogeneity, there has been some attempt to increase the communication efficiency under the existence of slow or straggler devices. Some works including \citep{damaskinos2020fleet} attempted to devise an asynchronous scheme. However, the staleness effect of slow devices might reduce the convergence time or even prevent it from reaching a stable point. Some other researches such as \citep{chen2021communication} provided a device selection scheme where only devices with the most contribution are chosen to reduce the number of communication round, while another group focused on the network resource management \citep{yang2020energy,chen2020convergence}. While the salient point of these researches is on the premise that there exist millions of devices (cross-device FL), there only exist a few studies on how to have a communication-efficient FL in a cross-silo setting where there is often a full participation in the aggregation~\citep{zhang2020batchcrypt}. In this case, due to the disparate resources and performance capabilities of devices, there will be a desynchronization in each round of communication, i.e., the more the heterogeneity, the more the desynchronization. This also poses a large latency resulting in a slow training procedure. Thus, an optimization scheme is required to reduce such delay in each communication round. 
\subsection{Contributions}
\textcolor{black}{Motivated by the edge devices' underlying relationship and their heterogeneous behavior, we introduce a novel framework, called Graph Federated Internet of Things Learning (GFIoTL).} The key contributions of the proposed framework are summarized as follows:

$\bullet$ The proposed GFIoTL approach incorporates a graph filtering aggregation rule based on the GSP concept dubbed ``G-Fedfilt'' where it opens the door to a whole new level of aggregation using graph filter design while incorporating FedAvg in a special case. To the best of our knowledge, this is the first work where graph filter design is used as an aggregation rule in FL which could potentially increase the devices' collaboration in training the models specific to each device. 

$\bullet$ G-Fedfilt brings personalization into FL algorithms while keeping a flow of information and unlike other personalized FL, involves edge device relationship on the graph. The proposed aggregation rule can also cluster edge devices inherently based on the graph's connectivity.

$\bullet$ \textcolor{black}{We propose a solution to system heterogeneity among CIoT devices in a cross-silo setting. The idea is to adaptively adjust the number of data samples and sparsity of models' gradients to dynamically synchronize the aggregation update. Additionally, the procedure is executed on the edge server based on the computational capabilities and the network connectivity of edge devices.}

The rest of this paper is organized as follows. In Section \ref{Se:problem}, the model description is presented. Section \ref{Se:GFL} introduces the main contributions of this work including the GFIoTL framework along with the parameter optimization scheme at the end. The experimental results are presented in Section \ref{Se:Exp} and finally, the conclusion with some remarks are stated in Section \ref{Se:conclusion}.
\section{Problem Description} \label{Se:problem}
In this section, we describe the backbone of the proposed framework in detail. Our description mainly focuses on a building with multiple \textcolor{black}{smart homes} where there exist various heterogeneous CIoT devices inside \textcolor{black}{each one} as depicted in Fig. \ref{fig:HidFL}. Although designed for a smart building, the model can be readily interpreted as a general-purpose framework for other fields of intelligent environment and devices such as smart factories, smart cities, and smart indoor environments \citep{sheikholeslami2021sub}. \textcolor{black}{Tables \ref{tab:notations} and \ref{tab:notations_2} provide the summary of the notations used hereafter.}
\begin{table}[!b]
\centering
\caption{\textcolor{black}{List of general notations used in this paper.}}
\label{tab:notations}
\begin{tabular}{|c|c|}
\hline
\textbf{Parameter} & \textbf{Description} \\ \hline
$A_i \times B_i \times C_i$ & Size of each room $c_j \in \mathcal{N}$ \\ \hline
$\mathcal{N}$ & Set of smart rooms \\ \hline
$N$ & Number of rooms \\ \hline
$\mathcal{K}_{c_j}$ & Set of devices in $c_j \in \mathcal{N}$ \\ \hline
$K_{c_j}$ & Number of devices in each room \\ \hline
$s_{i,c_j}$ &Device $i$ in room $c_j$ \\ \hline
$K$ & Total number of devices \\ \hline
$\mathcal{G}$ & Network graph between devices \\ \hline
$\mathcal{V}$ & Set of vertices of graph $\mathcal{G}$ \\ \hline
$\mathcal{A}_{s_{i,c_j}}$ & Neighbors' set of device $i$ in room $c_j$ \\ \hline
$\mathcal{E}$ & Set of edges of graph $\mathcal{G}$ \\ \hline
$\boldsymbol{A}$ & Adjacency matrix of graph $\mathcal{G}$ \\ \hline
$\boldsymbol{D}$ & Diagonal degree matrix of $\mathcal{G}$ \\ \hline
$\boldsymbol{L}$ & Combinatorial graph Laplacian of $\mathcal{G}$ \\ \hline
$\mathcal{D}_{s_{i,c_j}}$ & Dataset of device $i$ in room $c_j$ \\ \hline
$\mathcal{X}$ & Training data set \\ \hline
$\mathcal{Y}$ & Label set \\ \hline
$F_{s_{i,c_j}}(\cdot)$ & ML model for device $i$ in room $c_j$ \\ \hline
$\kappa_{s_{i,c_j}}$ & Associated weight of device $i$ in room $c_j$ \\ \hline
$L(\cdot)$ & Data sample loss function\\ \hline
$\tilde{L}(\cdot)$ & Local loss function \\ \hline
$L^\textrm{g}(\cdot)$ & Collaborative loss function \\ \hline
$\boldsymbol{g}_{s_{i,c_j}}$ & Gradient update of device $i$ in room $c_j$ \\ \hline
$R$ & Total number of comm. rounds \\ \hline
$\alpha_{s_{i,c_j}}$ & Number of local training at each comm. round \\ \hline
$\boldsymbol{G}$ & Matrix of gradient updates of all devices \\ \hline
$\boldsymbol{V}$ & Matrix of eigenvectors of $\boldsymbol{L}$ \\ \hline
$(v_i,\lambda_i)$ & $i^{th}$ eigenvector and eigenvalue of $\boldsymbol{L}$ \\ \hline
$\boldsymbol{\Lambda}$ & Diagonal matrix of eigenvalues of $\boldsymbol{L}$ \\ \hline

\end{tabular}
\end{table}

\textbf{Environment Specifications:}
\textcolor{black}{In this work, we consider a smart building consisting of $N$ smart homes each equipped with a Femto Access Point (FAP). Each smart home includes $K_{c_j}$ smart devices, indexed by $\mathcal{K}_{c_j}=\{s_{1,c_j},...,s_{K_{c_j},c_j} \}$, and with the total number of $K=\sum_{j=1}^{N}K_{c_j}$ devices. For simplicity and without loss of generality, we assume all devices belonging to a smart home are inside a single room. Each room, also called a cluster, is indexed by $c_j \in \mathcal{N}=[c_1,...,c_N]$ and has the size $A_i \times B_i \times C_i$ , $i \in 1,...,N$.}  We opt to select devices that are prevalent in real-world scenarios and mostly present in every common living room. To illustrate a more realistic scenario, consider each room having a set of common CIoT devices such as laptops, smartphones, smartwatches, tablets, and a smart TV creating a heterogeneous setup. In this case, each device $s_{i,c_j}\in \mathcal{K}_{c_j}$ is capable of performing a complex computation while having different specifications. 

\textbf{Graph Network:} \label{se:Graph}
Each device $s_{i,c_j}\in \mathcal{K}_{c_j}$ is connected to its neighbors in the subset $\mathcal{A}_{s_{i,c_j}}$. It is assumed that devices in the building constitute a bidirectional graph, denoted by $\mathcal{G}(\mathcal{V},\mathcal{E})$, where  $\mathcal{V}= \cup_{j=1}^{N} \mathcal{K}_{c_j}$ consists of all devices as the vertices of the graph and $\mathcal{E}= \cup_{j=1}^{N} \cup_{i=1}^{K_{c_j}} \mathcal{A}_{s_{i,c_j}}$ indicates all the edges connecting the devices. Moreover, the weight of the edges is indicated in the adjacency matrix $\boldsymbol{A}$, where $(\boldsymbol{A})_{ij}$ is the connection weight of devices $i$ and $j$. The entry $(\boldsymbol{A})_{ij}$ is chosen as either ``one'' or ``zero'' indicating the connectivity of devices $i$ and $j$ which is specified by
\begin{equation}
(\boldsymbol{A})_{ij}=\left\{\begin{matrix}
 1 & \textrm{if}~d_{ij}<d_{max}\\ 
 0& \textrm{otherwise}
\end{matrix}\right.,
\end{equation}
where $d_{ij}$ represents the distance between devices $i$ and $j$, $d_{max}$ denotes the maximum distance considered for paired devices. In addition, all devices have access to a nearby edge server to where they can share information and exchange data. The edge server can be any high processing machine located in the building, such as a central computer. It should be noted that the point of considering the graph connectivity between devices is to exploit such relationship and flow of information among devices which will be elaborated on further in the paper.

\textbf{Federated Learning Task:} 
Each device $s_{i,c_j}\in \mathcal{K}_{c_j}$ in cluster $c_j$, has a local dataset, denoted by $\mathcal{D}_{s_{i,c_j}}$, where $\mathcal{D}_{s_{i,c_j}}$ and $\mathcal{D}_{s_{i^\prime,c_j}}$, $i\neq i^\prime$, might or not have shared data samples. In this scheme, a supervised training procedure is considered where $(\mathcal{X},\mathcal{Y}) \in \mathcal{D}_{s_{i,c_j}}$ denotes the training data set $\mathcal{X}=\{\boldsymbol{x}_n\}_{n=1}^M$, $\boldsymbol{x}_n \in \mathbb{R}^\mathtt{d}$ and its corresponding label set $\mathcal{Y}=\{\boldsymbol{y}_n\}_{n=1}^M$, $\boldsymbol{y}_n \in (0,1)$. Here, $M$ represents the total number of data samples in $\mathcal{D}_{s_{i,c_j}}$ and $\mathbb{R}^\mathtt{d}$ specifies the feature space of the input data $\boldsymbol{x}_i$. In addition, a personalized model $F_{s_{i,c_j}}(x_n;\boldsymbol{\omega}_{s_{i,c_j}})$ with the loss function $L(F_{s_{i,c_j}}(x_n,\boldsymbol{\omega}_{s_{i,c_j}}),y_n)$, distributed among devices in the set $\mathcal{K}_{c_j}$ where $\boldsymbol{\omega}_{s_{i,c_j}} \in \mathbb{R}^B$, represents the parameters to be trained. Hence, the loss function of device $s_{i,c_j}$ is 
\begin{equation}
\tilde{L}( \boldsymbol{\omega}_{s_{i,c_j}})=\dfrac{1}{D_{s_{i,c_j}}}\sum_{(x_n,y_n) \in \mathcal{D}_{s_{i,c_j}}}L(F_{s_{i,c_j}}(x_n,\boldsymbol{\omega}_{s_{i,c_j}}),y_n),
\end{equation} 
where $D_{s_{i,c_j}} = |\mathcal{D}_{s_{i,c_j}}|$ denotes the cardinality of the set $\mathcal{D}_{s_{i,c_j}}$. The collaborative loss function of the framework is calculated as
\begin{equation}
L^\textrm{g}( \boldsymbol{\omega})= \sum_{j=1}^{N} \sum_{i=1}^{K_{c_j}} \kappa_{s_{i,c_j}} \tilde{L}( \boldsymbol{\omega}_{s_{i,c_j}}),
\end{equation}
where $\kappa_{s_{i,c_j}}$ is the associated weight of each device $s_{i,c_j}\in \mathcal{K}_{c_j}$. 
 The main goal of the FL is to solve the following minimization problem:
\begin{equation}
 \boldsymbol{\omega}_{s_{i,c_j}}^{\textrm{opt}} = \arg~\min L^\textrm{g}(\boldsymbol{\omega}),
\end{equation}
where $\boldsymbol{\omega}_{s_{i,c_j}}^{\textrm{opt}}$ is the optimum parameter weights for the model $F_{s_{i,c_j}}(x_n,\boldsymbol{\omega}_{s_{i,c_j}})$. Note that in case of searching a global parameter, we have $\boldsymbol{\omega}_{i,c_{j}}^{\textrm{opt}}=\boldsymbol{\omega}^{\textrm{opt}}_0$, $\forall s_{i,c_j}$, where $\boldsymbol{\omega}^{\textrm{opt}}_0$ is the global parameter.

\textcolor{black}{\textbf{Remark 1:} In case the input dimensions of ML models are different, one can equalize the dimensions using Encoder-Decoder (En-De) architectures. To do so, a preprocessing En-De unit can be used before the ML model of each device  $s_{i,c_j}\in \mathcal{K}_{c_j}$. Although En-De units are fed with different input dimensions, they can produce fixed-size latent spaces. Finally, the ML models are fed by the acquired latent space features that have the same dimensions.}
\section{Graph Federated Internet of Things Learning} \label{Se:GFL}
In this section, we introduce a novel paradigm in constructing an aggregation rule based on the graph connectivity of devices by exploiting Graph Signal Processing (GSP). We show that our new paradigm can incorporate FedAvg based on GSP under a certain condition which ultimately indicates the generality of the proposed approach. More specifically, we design a graph filter to not only aggregate the devices' gradients but also keep a level of personalization among devices. As another intrinsic capability of graph representation, we show that our approach can inherently group devices into different clusters based on their connections while allowing the flow of information. Note that by personalization, we refer to training specialized models tailored for each device rather than a single globally-shared model \citep{tan2021towards}.
\subsection{GFIoTL Overview}
We consider a cross-silo situation where all devices participate in each round of training. In addition, each device has its own specific ID where the edge server can identify. We also define a discrete-time set $\mathcal{T}=\{1,...,R\}$ and $t \in \mathcal{T}$ as the index time of the whole training procedure. Based upon the time notations, a fixed number of device-based updates occur before each aggregation time. After locally training each device $s_{i,c_j}\in \mathcal{K}_{c_j}$ for $\alpha_{s_{i,c_j}}$ epochs, the updated local weights $\boldsymbol{\omega}_{s_{i,c_j}}$ are sent to the edge server for the aggregation. Once the edge server collects enough model weights from edge devices, it performs model aggregation and sends the updated weights to specific devices. Although FedAvg is one of the most successful aggregation rules in FL, it does not consider any personalization due to its inherited averaging characteristic. In other words, with considering each device to have a specific domain due to data and label heterogeneity, FedAvg primarily acts as a domain-agnostic approach and ignores the domain specifics. 

To better handle the domain specificity in GFIoTL, we exploit $\mathcal{G}(\mathcal{V},\mathcal{E})$ and its representative adjacency matrix $\boldsymbol{A}$ further in the proposed paradigm. Moreover, it is more convenient to express the gradients update in a matrix form, denoted by $\boldsymbol{G}^{(t)} \in \mathbb{R}^{K \times B}$, where $i^{th}$ row of $\boldsymbol{G}^{(t)}$ corresponds to device $s_{i,c_j}$'s gradient update $\boldsymbol{g}^{(t)}_{s_{i,c_j}}=\boldsymbol{\omega}^{(t)}_{s_{i,c_j}} - \boldsymbol{\omega}^{(t-1)}_{s_{i,c_j}}$. In what follows, we will discuss some preliminaries regarding graphs in the Fourier domain as our aggregation rule is heavily dependent upon spatial frequencies and eigenfunctions.
\subsection{GSP Background} 
One of the important properties of a bidirectional graph is the so called \textit{combinatorial graph Laplacian}, defined as $\boldsymbol{L}=\boldsymbol{D}-\boldsymbol{A}$, where $\boldsymbol{D}$ denotes the diagonal degree matrix of $\mathcal{G}$ with $(\boldsymbol{D})_{ii}=\sum_{j=1}^{K}(\boldsymbol{A})_{ij}$. It can be shown that for a bidirectional graph such as $\mathcal{G}$, $\boldsymbol{L}$ is a positive semidefinite and all eigenvalues are non-negative real valued \citep{ortega2018graph}. Hence, it is common to choose the eigenvectors of the graph Laplacian $\boldsymbol{L}$ as the eigenfunctions, i.e, the basis of the Fourier transform. Consequently, the Fourier bases are extracted by the eigenvalue decomposition of the graph Laplacian as 
\begin{equation}
\boldsymbol{L} = \boldsymbol{V} \boldsymbol{\Lambda} \boldsymbol{V}^{T},
\end{equation}
where $\boldsymbol{V}=[\boldsymbol{v}_0,...,\boldsymbol{v}_{K-1}]$ represents the matrix of the $K$ eigenvectors of $\boldsymbol{L}$ and $\boldsymbol{\Lambda}=\textrm{diag}[\lambda_0,...,\lambda_{K-1}]$ is the corresponding eigenvalues. Thus, assuming eigenvalues are ordered based on the Total Variation (TV) of their eigenvectors used in \citep{ortega2018graph}, the corresponding eigenvector of $\lambda_0$ represents the DC basis and other eigenvectors of higher $\lambda_i$, $i=1,...,K-1,$ indicate the higher frequency bases of that particular graph. Consequently, the Graph Fourier Transform (GFT) and the Inverse GF Transform (IGFT) of the gradient matrix, denoted by $f(\boldsymbol{G}^{(t)})$ and $f^{-1}(f(\boldsymbol{G}^{(t)}))$, are calculated respectively as 
\begin{equation}
\boldsymbol{G}^{(t)}_f=f(\boldsymbol{G}^{(t)})= \boldsymbol{V}^{T}\boldsymbol{G}^{(t)},
\end{equation}
\begin{equation}
\boldsymbol{G}^{(t)}=f^{-1}(f(\boldsymbol{G}^{(t)})) = \boldsymbol{V}\boldsymbol{G}^{(t)}_f.
\end{equation}
Having the gradient update frequency coefficients $\boldsymbol{G}^{(t)}_f$ allows filtering the specific graph frequency by means of multiplication with the filter frequency response. Thus, there are three steps in graph filtering enumerated as $i)$ GFT, $ii)$ multiplication of the coefficient with the filter frequency response, and $iii)$ IGFT of the resultant. To achieve such filtering, a graph filter is defined in a matrix form, $\boldsymbol{H}$, such that $\boldsymbol{H}=\boldsymbol{V} h_s(\boldsymbol{\Lambda}) \boldsymbol{V}^{T}$, where $h_s(\boldsymbol{\Lambda})$ is the filter operator, i.e., $h_s(\boldsymbol{\Lambda})=\textrm{diag}[h_s(\lambda_0),...,h_s(\lambda_{k-1})]$. Hence, the filtered gradient updates is calculated in a compact form of multiplication as
\begin{align}\label{eq:filtering}
\widehat{\boldsymbol{G}}^{(t)} = \boldsymbol{H}\boldsymbol{G}^{(t)}&=\boldsymbol{\boldsymbol{V}}h_s(\boldsymbol{\Lambda}) \underbrace{ \boldsymbol{V}^{T}\boldsymbol{G}^{(t)}}_{\text{GFT}}\\ \label{eq:hsGf}
&=\boldsymbol{\boldsymbol{V}}\underbrace{h_s(\boldsymbol{\Lambda})\boldsymbol{G}^{(t)}_f}_{\text{ freq. response mul.}}\\\nonumber
&=\underbrace{\boldsymbol{\boldsymbol{V}}\widehat{\boldsymbol{G}}^{(t)}_f}_{\text{IGFT}}.
\end{align}
\subsection{GFIoTL Updates}
In this subsection, the two primary gradient updates required to achieve $\boldsymbol{\omega}^{\textrm{opt}}_{s_{i,c_j}}$ are discussed in detail.

\textbf{Local Model Update:} 
Each device  $s_{i,c_j}\in \mathcal{K}_{c_j}$
 performs a number of local updates independently on their dataset $\mathcal{D}_{s_{i,c_j}}$ to update their model parameters $\boldsymbol{\omega}_{s_{i,c_j}}$. Considering $\beta_{s_{i,c_j}}\subset \mathcal{D}_{s_{i,c_j}}$ as the mini-batch on which device $s_{i,c_j}$ is to be trained, the gradient estimate of the local update at time slot $t-1$ is calculated as 
\begin{equation}
\boldsymbol{\Delta}_{s_{i,c_j}}^{(t-1)}=\dfrac{1}{|\beta^{(t-1)}_{s_{i,c_j}}|}\sum_{(x_n,y_n) \in \beta^{(t-1)}_{s_{i,c_j}}} \bigtriangledown L(F_{s_{i,c_j}}(x_n,\boldsymbol{\omega}_{s_{i,c_j}}^{(t-1)}),y_n).
\end{equation} 

Therefore, the updated model parameter at time slot $t$ can be defined as 
\begin{equation}
\boldsymbol{\omega}_{s_{i,c_j}}^{(t)} = \boldsymbol{\omega}_{s_{i,c_j}}^{(t-1)} - \eta^{(t-1)}_{s_{i,c_j}}\boldsymbol{\Delta}_{s_{i,c_j}}^{(t-1)},~~\forall s_{i,c_j} \in \mathcal{K}_{c_j}  
\end{equation}
where $\eta^{(t-1)}_{s_{i,c_j}}$ indicates the local learning rate at time slot $t-1$.

\textbf{Global Aggregation Via Graph Filtering:}
After all model gradient updates of edge devices were collected by the edge server, it performs the aggregation process using the Graph Federated filtering (G-Fedfilt) as 
\begin{equation} \label{eq:G-Fedavg}
\boldsymbol{\widehat{G}}^{(t)}= \boldsymbol{H}\textrm{diag}[\kappa_1,...,\kappa_K]\boldsymbol{G}^{(t)},
\end{equation}
where $\textrm{diag}[\kappa_1,...,\kappa_K]$ represents the diagonal matrix of the weights associated to all edge devices and $\boldsymbol{H}$ is the graph filter to be designed based on the eigenvectors of $\boldsymbol{L}$ and the graph filter operator $h_s(\cdot)$.

 An interesting point in representing the connectivity of devices in graphs is that the FedAvg aggregation rule can be realized by applying a low-pass filter, denoted by $\boldsymbol{H}_{DC}$, with the cut-off frequency $f_{high}$ where $\lambda_1>f_{high}>\lambda_0$.
 In this case, the mean value is primarily obtained and broadcasted to all devices and therefore, there is no personalization for individual devices. On the other hand, consider an all-pass filter $\boldsymbol{H}_{all}$ with $f_{high}>\lambda_{K-1}$. Obviously, the resultant filtering of $\boldsymbol{G}^{(t)}$ is the matrix itself without any change, i.e., $\boldsymbol{G}^{(t)} = \boldsymbol{H}_{all}\boldsymbol{G}^{(t)}$. This outcome can be interpreted as a situation where only the data for each domain is considered and there is no flow of information among devices. In other words, we have a full personalization when aggregating the gradients using an all-pass filter. We can now make a rather concrete statement; \textit{as the cut-off frequency of the low-pass filter increases, so does the domain-specific behavior of the aggregation rule. Conversely, as the cut-off frequency decreases, so does the domain-agnostic of the aggregation rule.}
 
  Further elaboration on the mechanics of G-Fedfilt is presented in Appendix \ref{FirstAppendix}.
\subsection{Graph Filter Design}\label{se:GF}
\begin{figure}[!tb]
\centerline{\includegraphics[scale=.31]{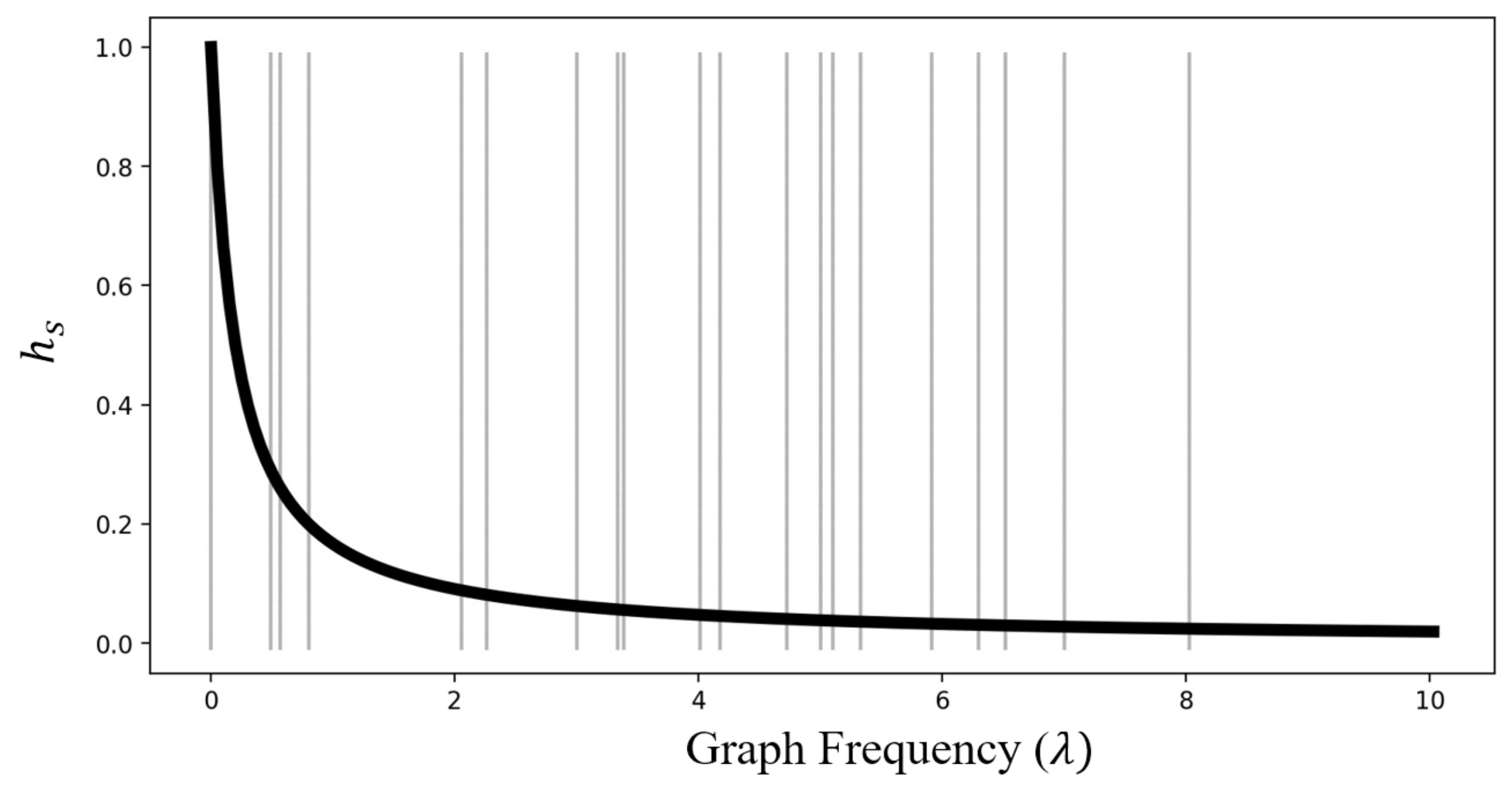}}
\caption{The graph frequency response of the proposed filter in G-Fedfilt for $\mu_{s}=5$. Horizontal lines indicate eigenvalues/graph frequencies.} 
\label{fig:GF}
\end{figure}

The underlying concept of using graph filters in G-Fedfilt is to consider both the domain-specific and the domain-agnostic gradient updates; to have a tunable personalization in FL. In this regard, we propose a graph filter operator in the graph frequency domain, behaving as a low-pass filter, that could achieve such a goal which is expressed as follows:
\begin{equation}
h_{s}(\lambda;\mu_{s})=\dfrac{1}{(1+\mu_{s}\lambda)},
\end{equation}
where $\mu_{s}$ is the tunable parameter of the filter. Therefore, given the eigenvalue $\lambda_i,~i=0,...,K-1$, the role of  $h_s(\lambda_i)$ is to attenuate the coefficients in $\boldsymbol{G}_f^{(t)}$ associated with higher frequencies while retaining the coefficients related to lower frequencies. This can be accomplished by the multiplication $h_s(\boldsymbol{\Lambda})\boldsymbol{G}^{(t)}_f$.
An illustration of such a filter is shown in Fig. \ref{fig:GF}. As seen, $h_s$ has a smooth transient low-pass filter which eventually goes to zero. It is worthwhile to remind that graph frequencies are represented by the eigenvalues of the graph Laplacian and therefore, vary with respect to $\boldsymbol{A}$. Here in Fig. \ref{fig:GF}, we considered $\boldsymbol{A}$ to be a $20\times 20$ symmetric matrix and the silver-colored horizontal lines show the eigenvalues of such matrix, i.e., graph frequencies ordered from zero to the highest eigenvalue corresponding to the highest graph frequency. Note that, the filter $h_s$ is not ideal and in fact, there are many ways to design a more suitable filter to be used in G-Fedfilt which opens up lots of rooms to design graph filters for the purpose of better aggregation in the GFIoTL.
 \begin{figure*}[!th]
\centerline{\includegraphics[scale=.45]{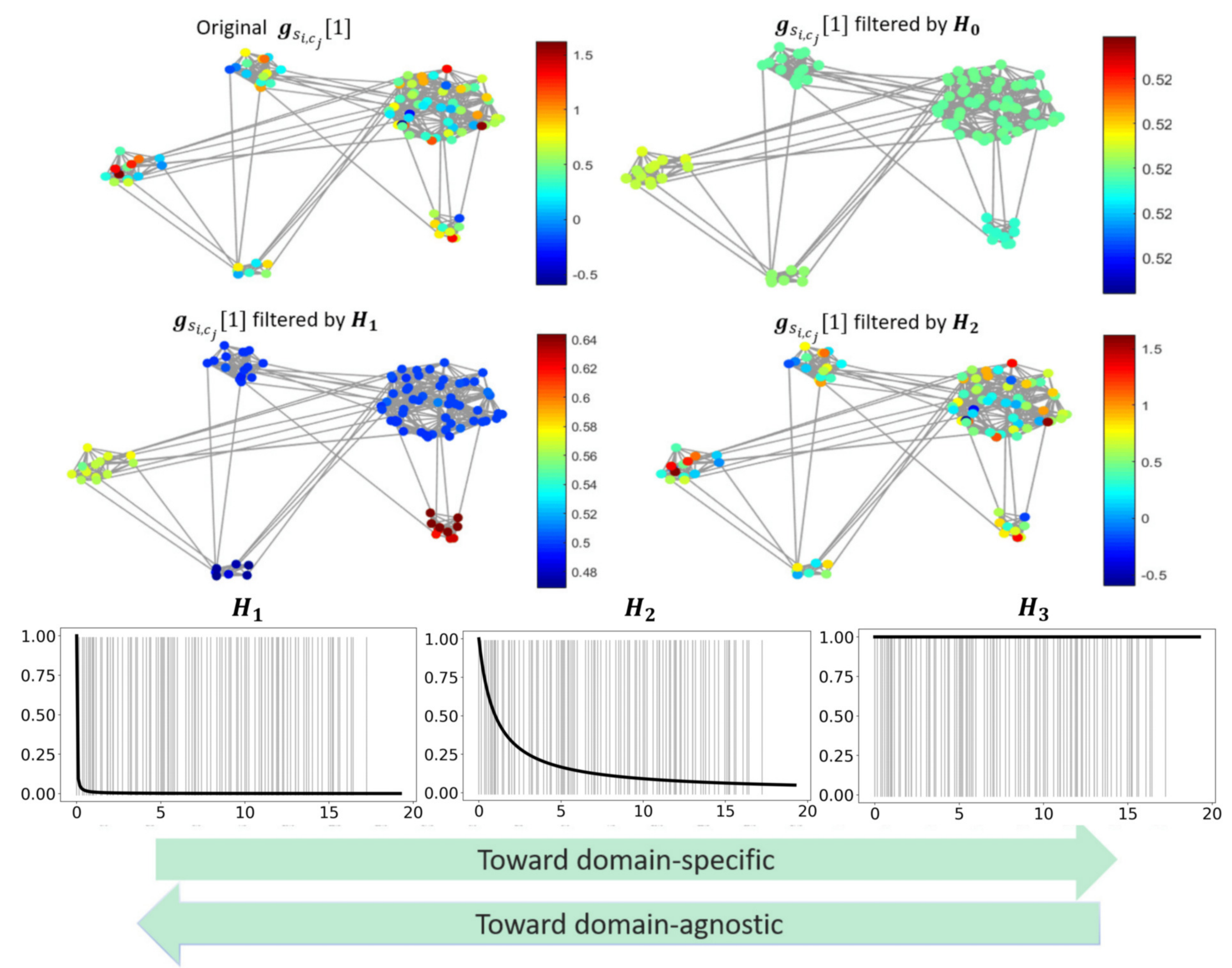}}
\caption{\textcolor{black}{An illustration of different graph filtering on a single shared gradient update $\boldsymbol{g}_{s_{i,c_j}}[1]$. Each node represents a device $s_{i,c_j}$ connected by $\mathcal{G}$.}}
\label{fig:Gfiltering}
\end{figure*}

 To have a better understanding of a low-pass graph filtering application, a graph signal along with its corresponding filtered versions is illustrated in Fig. \ref{fig:Gfiltering}. We used the publicly available PyGSP toolbox for this simulation~\citep{defferrard2017pygsp}. Here, there exist $100$ nodes corresponding to $100$ devices interconnected via a graph as a simulation environment of a building with multiple \textcolor{black}{smart rooms}. Each device $s_{i,c_j}$ is expressed with a color representing the first gradient element $\boldsymbol{g}_{s_{i,c_j}}[1]$ shared with all devices.  From this figure, it is seen that when the graph is filtered with $\boldsymbol{H}_{0}$, corresponding to $h_s(\lambda,10^{4})$, the average of the gradients will be calculated whereas, when filtered with $\boldsymbol{H}_{1}$, they have different aggregated values, though, they are still very close to the average. This behavior indicates that there is a flow of information among devices, and there also exists a level of personalization that can be smoothly controlled by adjusting the tunable parameter, i.e., $\mu_{s}$. Another salient point worth discussing is that when the graph is filtered by $\boldsymbol{H}_{1}$, each section of the graph appears to have a certain aggregated gradient value; as if each section turned into a cluster where the gradient value assigned to the nodes is the average value of that particular cluster. This behavior suggests that graph filtering can indeed cluster devices dynamically and intrinsically solely based on the graph's connectivity.  In the end, a filter such as $\boldsymbol{H}_2$ behaves as an all-pass filter where the local updates will be assigned as the next gradient updated without considering the neighboring gradients. The proposed G-Fedfilt algorithm is shown in Algorithm \ref{Al:GFed}.
\begin{table}[!tb]
\caption{\textcolor{black}{List of notations used in the proposed algorithms.}}
\centering
\label{tab:notations_2}
\begin{tabular}{|c|c|}
\hline
\textbf{Parameter} & \textbf{Description} \\ \hline
$d_{ij}$ & Distance between devices $i$ and $j$ \\ \hline
$d_{max}$ & Maximum distance between paired devices \\ \hline
$h_s(\cdot,\mu_s)$ & Graph filter operator \\ \hline
$\mu_s$ & Tunable parameter of $h_s$ \\ \hline
$\boldsymbol{\widehat{G}}$ & Updated gradient matrix \\ \hline
$\boldsymbol{H}$ & Graph filter in matrix form \\ \hline
$\mathcal{B}_{s_{i,c_j}}$ & Mini-batch set for device $i$ in room $c_j$ \\ \hline
$\beta_{s_{i,c_j}}$ & Mini-batch in the set $\mathcal{B}_{s_{i,c_j}}$ \\ \hline
$\boldsymbol{\Delta}_{s_{i,c_j}}$ & Local gradient after each local update \\ \hline
$\boldsymbol{\omega}_{s_{i,c_j}}$ & Model weights for device $i$ in room $c_j$ \\ \hline
$\eta_{s_{i,c_j}}$ & Local learning rate \\ \hline
$z_{s_{i,c_j}}$ & Sparsification controlling parameter \\ \hline
$q_{s_{i,c_j}}$ & Local data number controlling parameter \\ \hline
$Th_{s_{i,c_j}}$ & Sparsification threshold \\ \hline
$T$ & Deadline for each Comm. round \\ \hline
$\rho_{s_{i,c_j}}$ & Computing intensity \\ \hline
$f_{s_{i,c_j}}$ & Number of CPU cycles per sec.\\ \hline
$\varsigma$ & Effective switch capacitance\\ \hline
$b_{s_{i,c_j}}$ & Allocated bandwidth\\ \hline
$p^{\textrm{tran}}_{s_{i,c_j}}$ & Average transmission power\\ \hline
$n_0$ & Power spectral density of the Gaussian noise\\ \hline
$r_{s_{i,c_j}}$ & Transmission rate\\ \hline
$(E^{\textrm{comp}}_{s_{i,c_j}},E^{\textrm{tran}}_{s_{i,c_j}})$ & Comp. and comm. energy consumption \\ \hline
$(\mu_1,\mu_2,\mu_3)$ &  Problem $\mathcal{P}^{\prime \prime} $  optimization coefficients \\ \hline
$\alpha_{s_{i,c_j}}$ & Number of local training at each comm. round \\ \hline
$\chi(\varphi(\cdot;z_{s_{i,c_j}}))$ & Data size of the sparsification function \\ \hline
$\psi(\cdot;q_{s_{i,c_j}})$ & Number of data samples used for training \\ \hline
$\chi(\varphi(\cdot;z_{s_{i,c_j}}))$ & Data size of the sparsification function \\ \hline
$\psi(\cdot;q_{s_{i,c_j}})$ & Number of data samples used for training \\ \hline
$(\tau^{\textrm{comp}}_{s_{i,c_j}},\tau^{\textrm{tran}}_{s_{i,c_j}})$ & Computation and Comm. time \\ \hline
\end{tabular}
\end{table}
\begin{algorithm}[!t]
\caption{The proposed G-Fedfilt Algorithm}\label{Al:GFed}
\hspace*{\algorithmicindent} \textbf{Input:} model weights $\boldsymbol{\omega}^{(0)}_{s_{i,c_j}}$,$N$, $K_{c_j}$, $K$, number of comm. rounds $R$, number of local training updates $\alpha_{s_{i,c_j}}$, device distances $d_{ij}$, $d_{max}$, filter parameter $\mu_s$. \\
 
\begin{algorithmic}[1]
\Procedure{G-Fedfilt}{}
\BState \emph{Edge Server Initialization:}
\State $(\boldsymbol{A})_{ij}\gets \left\{\begin{matrix}
1 & \textrm{if}~d_{ij}<d_{max}\\ 
 0& \textrm{otherwise}
\end{matrix}\right.,$
\State $\kappa_{s_{i,c_j}}\gets \dfrac{|D_{s_{i,c_j}}|}{\sum^{N}_{j=1}\sum^{K_{c_j}}_{i=1}|D_{s_{i,c_j}}|}$
\State Get $\boldsymbol{\Lambda}$ and $\boldsymbol{V}$ by eigenvalue decomp. of $\boldsymbol{L}=\boldsymbol{D}-\boldsymbol{A}$
\State $h_s(\boldsymbol{\Lambda};\mu_s)\gets h_s(\lambda;\mu_s)$ \Comment {Matrix format}
\State $\boldsymbol{H}\gets \boldsymbol{\boldsymbol{V}} h_s(\boldsymbol{\Lambda}) \boldsymbol{V}^{T}$ 
\For {each round $t=1,...,R$}
		\BState \emph{Edge user Side}
		\For{each device $s_{i,c_j}\in \mathcal{K}_{c_j}$ \textbf{in parallel} }
				
			\State$\boldsymbol{\omega}^{(t+1)}_{s_{i,c_j}} \gets$ \textsc{ClientUpdate($\boldsymbol{\omega}^{(t)}_{s_{i,c_j}}$,$\alpha_{s_{i,c_j}}$)}
			\State $\boldsymbol{g}^{(t+1)}_{s_{i,c_j}}\gets \boldsymbol{\omega}^{(t+1)}_{s_{i,c_j}}-\boldsymbol{\omega}^{(t)}_{s_{i,c_j}} $		
			\State Transmit $\boldsymbol{g}^{(t+1)}_{s_{i,c_j}}$ to the edge server
		\EndFor{\textbf{End for}}
		\BState \emph{Edge Server Side}: 
		\State Stack $\boldsymbol{g}^{(t+1)}_{s_{i,c_j}}$ to create $\boldsymbol{G}^{(t+1)}$
		\State $\boldsymbol{\widehat{G}}^{(t+1)} \gets \boldsymbol{H}\textrm{diag}[\kappa_1,...,\kappa_K]\boldsymbol{G}^{(t+1)},$
		\State Broadcast $\boldsymbol{\widehat{G}}^{(t+1)}$
	\EndFor{\textbf{End for}}	
\EndProcedure{\textbf{End procedure}}

\end{algorithmic}
\begin{algorithmic}[1]
\Function{ClientUpdate}{$\boldsymbol{\omega}^{(t)}_{s_{i,c_j}}$,$\alpha_{s_{i,c_j}}$}

	\For{each local update $l=1,...,\alpha_{s_{i,c_j}}$}
		\State $\mathcal{B}_{s_{i,c_j}}\gets$ create a set of mini-batches from $\mathcal{D}_{s_{i,c_j}}$
		\For{each mini-batch $\beta_{s_{i,c_j}}\in \mathcal{B}_{s_{i,c_j}}$}
			\State $\boldsymbol{\Delta}_{s_{i,c_j}}^{(t-1)}\gets \dfrac{1}{|\beta^{(t)}_{s_{i,c_j}}|}\sum_{(x_n,y_n) \in \beta^{(t)}_{s_{i,c_j}}} \bigtriangledown L(F_{s_{i,c_j}},y_n)$
			\State $\boldsymbol{\omega}_{s_{i,c_j}}^{(t+1)} \gets \boldsymbol{\omega}_{s_{i,c_j}}^{(t)} - \eta^{(t)}_{s_{i,c_j}}\boldsymbol{\Delta}_{s_{i,c_j}}^{(t)}$
		\EndFor{\textbf{End for}}
	\EndFor{\textbf{End for}}
	\State \textbf{return $\boldsymbol{\omega}_{s_{i,c_j}}^{(t+1)}$}
\EndFunction{\textbf{End function}}
\end{algorithmic}
\end{algorithm}


\subsection{Parameter Optimization For Communication Efficiency} \label{se:opt}
The primary goal of this section is to reduce the latency and the desynchronization caused by the system heterogeneity of devices in each $c_j \in \mathcal{N}$. 
 In the following, we elaborate on the details and formulations of this problem and our solution embedded in the proposed GFIoTL.

We define a deadline $T$ in second for each round of communication within which all gradient updates must have already been received by the edge server. This means that each device $s_{i,c_j}\in \mathcal{K}_{c_j}$ in room $c_j$ must perform its local update computation and communication before $T$. This parameter depends on the index $t$ meaning that its value is subject to change at the onset of every first device-based update. This is due to the high-performance variability and heterogeneity, even for the same device \citep{nishio2019client}. Here, for simplicity of notation presentation, we drop the time index $t$ and make the point that each following calculation is performed at the beginning of the first device-based update back in the edge server.

There exist two sources of energy consumption in each device $s_{i,c_j}\in \mathcal{K}_{c_j}$. The first source is the computation energy caused by the local training and the second one is the communication energy wasted during the data transmission. For each, we calculate the time of which they perform the task (i.e., computation and communication) as demonstrated in \citep{cui2017software}.

\textbf{Local Computation:}
 Let $\rho_{s_{i,c_j}}$ denote the computing intensity (in the unit of CPU cycle per sample) of the device $s_{i,c_j}$ and $f_{s_{i,c_j}}$ the computation capacity measured by the number of CPU cycles per second. Therefore, the computation time for $\alpha_{s_{i,c_j}}$ rounds of local update for device $s_{i,c_j}$ is calculated as 
\begin{equation}\label{comp_delay}
\tau^{\textrm{comp}}_{s_{i,c_j}} = \dfrac{\alpha_{s_{i,c_j}} \psi(\mathcal{D}_{s_{i,c_j}};q_{s_{i,c_j}}) \rho_{s_{i,c_j}}}{f_{s_{i,c_j}}},~\forall s_{i,c_j} \in \mathcal{K}_{c_j},
\end{equation}
where $\psi(\mathcal{D}_{s_{i,c_j}};q_{s_{i,c_j}})=[q_{s_{i,c_j}}|\mathcal{D}_{s_{i,c_j}}|]$ represents a function characterized by  $q_{s_{i,c_j}} \in (0,1)$ for how much percentage of the number of local dataset is to be used in the training process at each communication round. 
Moreover, the total energy consumption due to the computation at each device-based update is derived as
\begin{equation}\label{comp_energy}
E_{s_{i,c_j}}^{\textrm{comp}} =\varsigma \alpha_{s_{i,c_j}}\psi(\mathcal{D}_{s_{i,c_j}};q_{s_{i,c_j}}) \rho_{s_{i,c_j}}f^2_{s_{i,c_j}},  
\end{equation} 
where $\varsigma$ represents the effective switched capacitance that depends on
the chip architecture.

\textbf{Wireless Communication:}
With regard to the communication between smart devices and the edge server, we consider an Orthogonal Frequency-Division Multiple Access (OFDMA) technique where a subset of subcarriers are assigned to the smart devices. Accordingly, denoting $b_{s_{i,c_j}}$ as the allocated bandwidth to device $s_{i,c_j}$, the achievable transmission rate of device $s_{i,c_j}$ is obtained as
\begin{equation}
r_{s_{i,c_j}}=b_{s_{i,c_j}}\log_2\left(1+\dfrac{\xi_{s_{i,c_j}}p^{\textrm{tran}}_{s_{i,c_j}}}{n_0b_{s_{i,c_j}}}\right),~~\forall s_{i,c_j} \in \mathcal{K}_{c_j},
\end{equation}
where $\xi_{s_{i,c_j}}$ denotes the channel gain between device $s_{i,c_j}$ and the edge server, $p^{\textrm{tran}}_{s_{i,c_j}}$ represents the average transmit power of device $s_{i,c_j}$, and $n_0$ is the power spectral density of the Gaussian noise. Since a limited bandwidth is available, we also have the constraint $\sum_{i=1}^{k_{c_j}}\sum_{j=1}^{N}b_{s_{i,c_j}}\leq \beta$ where $\beta$ is the total bandwidth. Moreover, we define $\chi(\varphi(\boldsymbol{g}_{s_{i,c_j}};z_{s_{i,c_j}})):\mathbb{R}^B \rightarrow \mathbb{R}$ to determine the data size of the $\varphi$'s output. Note that $\chi(\varphi(\boldsymbol{g}_{s_{i,c_j}};z_{s_{i,c_j}}))$ is increasing with respect to $z_{s_{i,c_j}}$. As an example, decreasing $z_{s_{i,c_j}}$ creates a sparser gradient output derived from $\varphi(\boldsymbol{g}_{s_{i,c_j}};z_{s_{i,c_j}})$ and consequently, the data size indicated by $\chi(\varphi(\boldsymbol{g}_{s_{i,c_j}};z_{s_{i,c_j}}))$ becomes smaller.
Thus, the transmission time between each device $s_{i,c_j}$ and the edge server is derived as follows:
\begin{equation}\label{comm_delay}
\tau^{\textrm{tran}}_{s_{i,c_j}} = \dfrac{ \chi(\varphi(\boldsymbol{g}_{s_{i,c_j}};z_{s_{i,c_j}}))}{r_{s_{i,c_j}}}.
\end{equation}
Accordingly, the energy consumption caused by the transmission is given by 
\begin{equation}\label{comm_energy}
E_{s_{i,c_j}}^{\textrm{tran}} =p^{\textrm{tran}}_{s_{i,c_j}} \tau^{\textrm{tran}}_{s_{i,c_j}}.  
\end{equation}
Furthermore, the latency of device $s_{i,c_j}$ at each communication round is defined as
\begin{equation}\label{total_delay}
\tau_{s_{i,c_j}}=\tau^{\textrm{comp}}_{s_{i,c_j}}+\tau^{\textrm{tran}}_{s_{i,c_j}} \leq T.
\end{equation}

 To reduce the total amount of communication and computation time for each device $s_{i,c_j}$, four parameters need to be tuned: $i)$ the number of rounds in each local update, $\alpha_{s_{i,c_j}}$, $ii)$ the portion of local data set used for training in each round, $q_{s_{i,c_j}}$, $iii)$ the sparsification parameter $z_{s_{i,c_j}}$, and $iv)$ the deadline in each communication round, $T$.
In this regard, we aim to minimize the latency in each communication round, $T$, while maximizing $\alpha_{s_{i,c_j}}$, $q_{s_{i,c_j}}$, and $z_{s_{i,c_j}}$. To achieve this goal, we formulate the following optimization problem ($\mathcal{P}$) which is solved by the edge server at the beginning of each round:
\begin{align}
\mathcal{P}: &\max_{\boldsymbol{\alpha},\boldsymbol{q},\boldsymbol{z},T} ~~~~~ \boldsymbol{F} = \{\alpha_{s_{i,c_j}},q_{s_{i,c_j}},z_{s_{i,c_j}},\frac{1}{T}\} \label{obj_fun}\\\nonumber
&\textbf{s.t.}\\
&(C_1)~\tau^{\textrm{comp}}_{s_{i,c_j}}+ \tau^{\textrm{tran}}_{s_{i,c_j}} \leq T~,~\forall s_{i,c_j} \in \mathcal{K}_{c_j},  \\
&(C_2)~E_{s_{i,c_j}}^{\textrm{comp}} +E_{s_{i,c_j}}^{\textrm{tran}}\leq E_{s_{i,c_j}}^{\textrm{max}}~,~\forall s_{i,c_j} \in \mathcal{K}_{c_j},\label{e_bound}  \\
&(C_3)~\alpha_{s_{i,c_j}}^{\textrm{min}} \leq \alpha_{s_{i,c_j}} \leq \alpha_{s_{i,c_j}}^{\textrm{max}}, \alpha_{s_{i,c_j}} \in \mathbb{N},~\forall s_{i,c_j} \in \mathcal{K}_{c_j}, \label{alpha_bound}\\
&(C_4)~q_{s_{i,c_j}}^{\textrm{min}} \leq q_{s_{i,c_j}}\leq 1~,~\forall s_{i,c_j} \in \mathcal{K}_{c_j}, \label{q_bound}\\
&(C_5)~z_{s_{i,c_j}}^{\textrm{min}} \leq z_{s_{i,c_j}}\leq 1~,~\forall s_{i,c_j} \in \mathcal{K}_{c_j}, \label{z_bound}\\\nonumber
\end{align}
where $\boldsymbol{\alpha}=[\alpha_{s_{1,c_1}},...,\alpha_{s_{K_{c_N},c_N}}]$, $\boldsymbol{q}=[q_{s_{1,c_1}},...,q_{s_{K_{c_N},c_N}}]$, and $\boldsymbol{z}=[z_{s_{1,c_1}},...,z_{s_{K_{c_N},c_N}}]$ are the vectors containing the variables of the optimization problem. As it can be seen from \eqref{obj_fun}, problem $\mathcal{P}$ is a multi-objective problem. Note that constraints \eqref{q_bound} and \eqref{z_bound} are linear and do not violate the convexity of the problem. However, functions $\psi(\mathcal{D}_{s_{i,c_j}};q_{s_{i,c_j}})$ and $\chi(\varphi(\boldsymbol{g}_{s_{i,c_j}};z_{s_{i,c_j}}))$ are non-convex with respect to $q_{s_{i,c_j}}$ and $z_{s_{i,c_j}}$, thus, according to \eqref{comp_delay} and \eqref{comm_delay}, constraint $(C_1)$ is non-convex for $q_{s_{i,c_j}}$ and $z_{s_{i,c_j}}$. Similarly, it can be understood from \eqref{comp_energy} and \eqref{comm_energy} that the left side of constraint $(C_2)$ provides non-convexity for variables $q_{s_{i,c_j}}$ and $z_{s_{i,c_j}}$. \textcolor{black}{Moreover, the integer parameter $\alpha_{s_{i,c_j}}$ used in constraints $(C_1)$-$(C_3)$ is non-convex.}

\begin{figure}[tbp]
\centerline{\includegraphics[scale=.4]{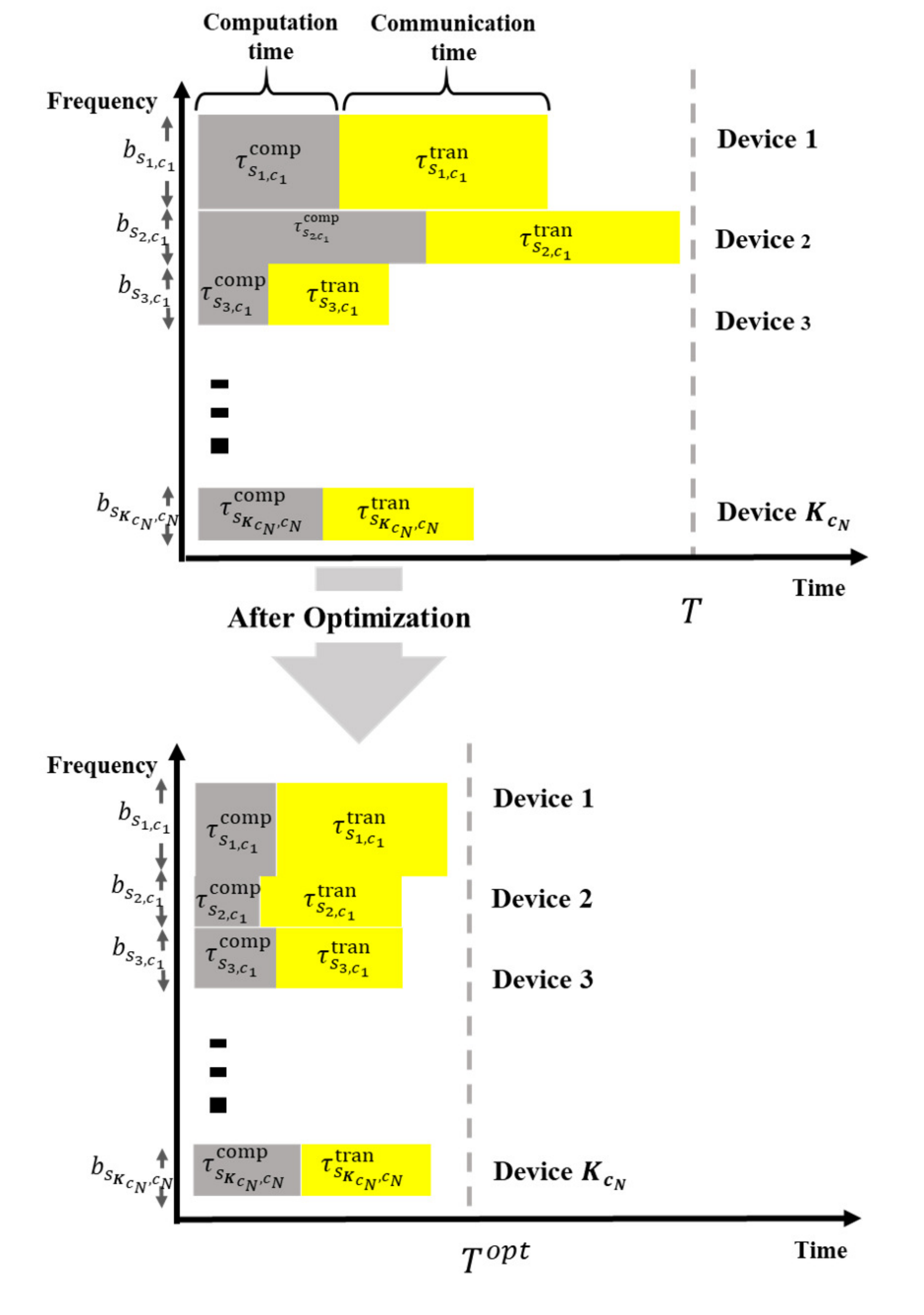}}
\caption{An illustration of the system heterogeneity in the training phase and the effect of the proposed parameter optimization scheme on reducing the latency in each communication round.}
\label{fig:timeplan}
\end{figure}

Additionally, the objective function and the constraints in optimization problem $\mathcal{P}$ are convex with respect to \textcolor{black}{$T$}. Hence, to solve optimization problem $\mathcal{P}$, we first derive the optimum value for $T$, denoted by $T^{\textrm{opt}}$. Then, given the optimal value of $T$, we substitute it in problem $\mathcal{P}$ and optimize the other optimization variables, i.e., $\boldsymbol{\alpha}$, $\boldsymbol{q}$ and $\boldsymbol{z}$. As it can be seen from Fig. \ref{fig:timeplan}, the minimum value of $T$ is equal to the latency of the device with the biggest delay. In other words, 
\begin{equation}\label{T_optimal}
	T = \text{max}~~~ \tau^{\textrm{comp}}_{s_{i,c_j}}+ \tau^{\textrm{tran}}_{s_{i,c_j}},~\forall s_{i,c_j} \in \mathcal{K}_{c_j} .
\end{equation}
Hence, it is aimed to minimize the latency of the slowest device participating in the current communication round. To reach the minimum value for $T$, we first derive a lower bound for the latency of each smart device. According to \eqref{comp_delay} and \eqref{comm_delay}, and considering that $\psi(.)$ 
and $\varphi(.)$ are increasing functions with respect to \textcolor{black}{$\alpha_{s_{i,c_j}}$,} $q_{s_{i,c_j}}$ and $z_{s_{i,c_j}}$, respectively, it is concluded that both terms of delays (i.e., transmission and computation terms) have an increasing relationship with the optimization variables, $\alpha_{s_{i,c_j}}$, $q_{s_{i,c_j}}$, and $z_{s_{i,c_j}}$. Therefore, the minimum latency for each device $s_{i,c_j}$ is achieved when the optimization variables are equal to their lowest values (i.e., $\widetilde{\alpha}_{s_{i,c_j}} = \alpha_{s_{i,c_j}}^\text{min}$, $\widetilde{q}_{s_{i,c_j}} = q_{s_{i,c_j}}^\text{min}$, and $\widetilde{z}_{s_{i,c_j}} = z_{s_{i,c_j}}^\text{min}$).
Thus, the lower bound for the latency of each smart device $s_{i,c_j}$, denoted by $\widetilde{\tau}_{s_{i,c_j}}$, is obtained by applying $\widetilde{\alpha}_{s_{i,c_j}}$, $\widetilde{q}_{s_{i,c_j}}$, and $\widetilde{z}_{s_{i,c_j}}$ in \eqref{total_delay}. Finally, according to \eqref{T_optimal}, the optimum value for $T$ is the latency of the slowest device given as
\begin{equation}\label{T_opt}
	T^{\textrm{opt}} = \text{max}~\widetilde{\tau}_{s_{i,c_j}}.
\end{equation}

To proceed with solving problem $\mathcal{P}$, we substitute $T=T^\text{opt}$ and relax constraints $(C_1)$ and $(C_2)$ which are non-convex for the optimization variables \textcolor{black}{$\alpha_{s_{i,c_j}}$, $q_{s_{i,c_j}}$ and $z_{s_{i,c_j}}$}. To cope with the non-convexity caused by $\psi(\mathcal{D}_{s_{i,c_j}};q_{i,c_j})$ in \eqref{comp_delay} and \eqref{comp_energy}, we use the following estimation in constraints $(C_1)$ and $(C_2)$ as $q_{s_{i,c_j}}|\mathcal{D}_{s_{i,c_j}}| >> 1$ is a relatively large number.
\begin{equation}\label{q_approx}
	\psi(\mathcal{D}_{s_{i,c_j}};q_{s_{i,c_j}})=[q_{s_{i,c_j}}|\mathcal{D}_{s_{i,c_j}}|]\approx q_{s_{i,c_j}}|\mathcal{D}_{s_{i,c_j}}|.
\end{equation}
To remove the non-convexity imposed by function $\varphi(.)$, it can be shown that $\chi(\varphi(\boldsymbol{g}_{s_{i,c_j}};z_{s_{i,c_j}}))$ is increasing with respect to $z_{s_{i,c_j}}$. In other words, maximizing $\chi(\varphi(\boldsymbol{g}_{s_{i,c_j}};z_{s_{i,c_j}}))$ leads to maximizing $z_{s_{i,c_j}}$. Thus, we replace $z_{s_{i,c_j}}$ in the objective function with $\chi(\varphi(\boldsymbol{g}_{s_{i,c_j}};z_{s_{i,c_j}}))$ which makes the both constraints $(C_1)$ and $(C_2)$ convex. \textcolor{black}{In addition, we relax the integer variable $\alpha_{s_{i,c_j}}$ to remove the non-convexity in constraints $(C_1)$-$(C_3)$. Finally,} by applying the optimal value of $T$ in \eqref{T_opt}, problem $\mathcal{P}$ is converted to the following optimization problem: 

\begin{align}
	\mathcal{P'}: &\max_{\boldsymbol{\alpha},\boldsymbol{q},\boldsymbol{\varphi}} ~~~ \widetilde{\boldsymbol{F}} = \{\alpha_{s_{i,c_j}}, q_{s_{i,c_j}},\chi(\varphi(\boldsymbol{g}_{s_{i,c_j}};z_{s_{i,c_j}}))\} \\\nonumber
	&\textbf{s.t.} \\ 
	&(C_1)~\tau^{\textrm{comp}}_{s_{i,c_j}}+ \tau^{\textrm{tran}}_{s_{i,c_j}} = T^\text{opt}~,~\forall s_{i,c_j} \in \mathcal{K}_{c_j} ,\label{T_bound_new} \\
	&(C_3)~\alpha_{s_{i,c_j}}^{\textrm{min}} \leq \alpha_{s_{i,c_j}} \leq \alpha_{s_{i,c_j}}^{\textrm{max}}~,~\forall s_{i,c_j} \in \mathcal{K}_{c_j}\label{a_bound_new},\\
	&(\ref{e_bound}),(\ref{q_bound}),(\ref{z_bound}). \nonumber
\end{align}

The next step in solving the optimization problem  $\mathcal{P}$ is to maximize $\alpha_{s_{i,c_j}}$, $q_{s_{i,c_j}}$, and $\chi(\varphi(\boldsymbol{g}_{s_{i,c_j}};z_{s_{i,c_j}}))$ according to the optimum value of $T= T^{\textrm{opt}}$ such that smart devices deliver their local models to the edge server at the same time. 
In order to solve $\mathcal{P}'$, we first make a single dimensionless objective function by dividing the objectives to their nominal values, $\alpha_{s_{i,c_j}}^{\textrm{max}}$, $q_{s_{i,c_j}}^{\textrm{max}}$, and $\chi_{s_{i,c_j}}^{\textrm{max}}$, respectively, for $\alpha_{s_{i,c_j}}$, $q_{s_{i,c_j}}$, and $\chi(\varphi(\boldsymbol{g}_{s_{i,c_j}};z_{s_{i,c_j}}))$. Additionally, we assign weight coefficients $\mu_1$, $\mu_2$, and $\mu_3$, where $\mu_1 + \mu_2 + \mu_3 = 1$, to model the relative importance among variables for each device. Thus, the optimization problem $\mathcal{P}'$ is converted to the following problem:
\begin{align}
	\mathcal{P}'': &\max_{\boldsymbol{\alpha},\boldsymbol{q},\boldsymbol{z}} ~~~ \frac{\mu_1}{\alpha_{s_{i,c_j}}^{\textrm{max}}}\alpha_{s_{i,c_j}} +  \frac{\mu_2}{q_{s_{i,c_j}}^{\textrm{max}}}q_{s_{i,c_j}}\\\nonumber
	 &+\frac{\mu_3}{\chi_{s_{i,c_j}}^{\textrm{max}}}\chi(\varphi(\boldsymbol{g}_{s_{i,c_j}};z_{s_{i,c_j}})) \\\nonumber
	&\textbf{s.t.} ~~~~ (\ref{e_bound}),(\ref{q_bound}),(\ref{z_bound}),(\ref{T_bound_new}),(\ref{a_bound_new}). 
\end{align}
Problem $\mathcal{P}''$ is convex and can be solved with the standard optimization techniques, such as convex optimization and barrier methods \citep{Boyd} that are employed in MATLAB CVX optimization toolbox \citep{grant2014cvx}. Note that the optimal value for $z_{s_{i,c_j}}$ can be obtained with the optimum $\chi(\varphi(\boldsymbol{g}_{s_{i,c_j}};z_{s_{i,c_j}}))$ as it is increasing with respect to $z_{s_{i,c_j}}$.
\section{Simulation Results} \label{Se:Exp}
In this section, we evaluate the proposed G-Fedfilt algorithm in the introduced GFIoTL framework and the proposed parameter optimization for communication efficiency in FL. \textcolor{black}{The cornerstone of the proposed GFIoTL framework is task-independent meaning that, given the ML models deployed on various devices, the proposed approach trains the models in a federated manner and based on their connections on the graph. In summary, there are four components required for the GFIoTL framework to perform on smart home applications such as HAR; $1)$ device hardware specifications for parameter optimization, $2)$ the ML models, $3)$ the data with which models are being trained, and $4)$ a criterion for the representative graph's connectivity. Based on such requirements, we test the applicability and performance of the proposed scheme in a typical simulation environment similar to smart home/room application scenarios. We also aim to find the answer to the following questions:}  

$\bullet$ Does G-Fedfilt incorporate FedAvg?

$\bullet$ What is the impact of graph filtering in model personalization under label and data heterogeneity?

$\bullet$ Can we personalize the models over their local datasets while keeping a level of generalization when it comes to data from other distributions?

$\bullet$ Does exploiting devices' relationship in the form of a graph contribute to the models' accuracy?

$\bullet$ Is it possible to decrease the communication round delay while involving devices with the system heterogeneity and performance variability.

To seek the answer to above questions and evaluate GFIoTL performance on the simulation environment, we conduct various numerical experiments with different scenarios. The specification of the simulation is as follows.

\subsection{Experimental Setup}
\textcolor{black}{Throughout the simulations, we consider $K=20$ heterogeneous edge devices spread out in $N=4$ smart rooms with the square area of $A_j \times B_j \times C_j=10^3$ m$^3$ such that $K_{c_j}\sim Du(4,7)$,  $\forall c_j$, where  $ Du(\cdot)$ indicates the discrete uniform distribution.} It is assumed that the edge server is close by, and the link between devices and the edge server is error-less. \textcolor{black}{It is worthwhile to note that although the purpose of the setup is for the training phase, one can exploit such an arrangement for the inference phase in the case of HAR applications. In this scenario, smart devices in each room collect data on the client and make an inference about their activity. Eventually, the edge server gathers the decisions and makes the final vote.}

\begin{figure}[tbp]
\centerline{\includegraphics[scale=.27]{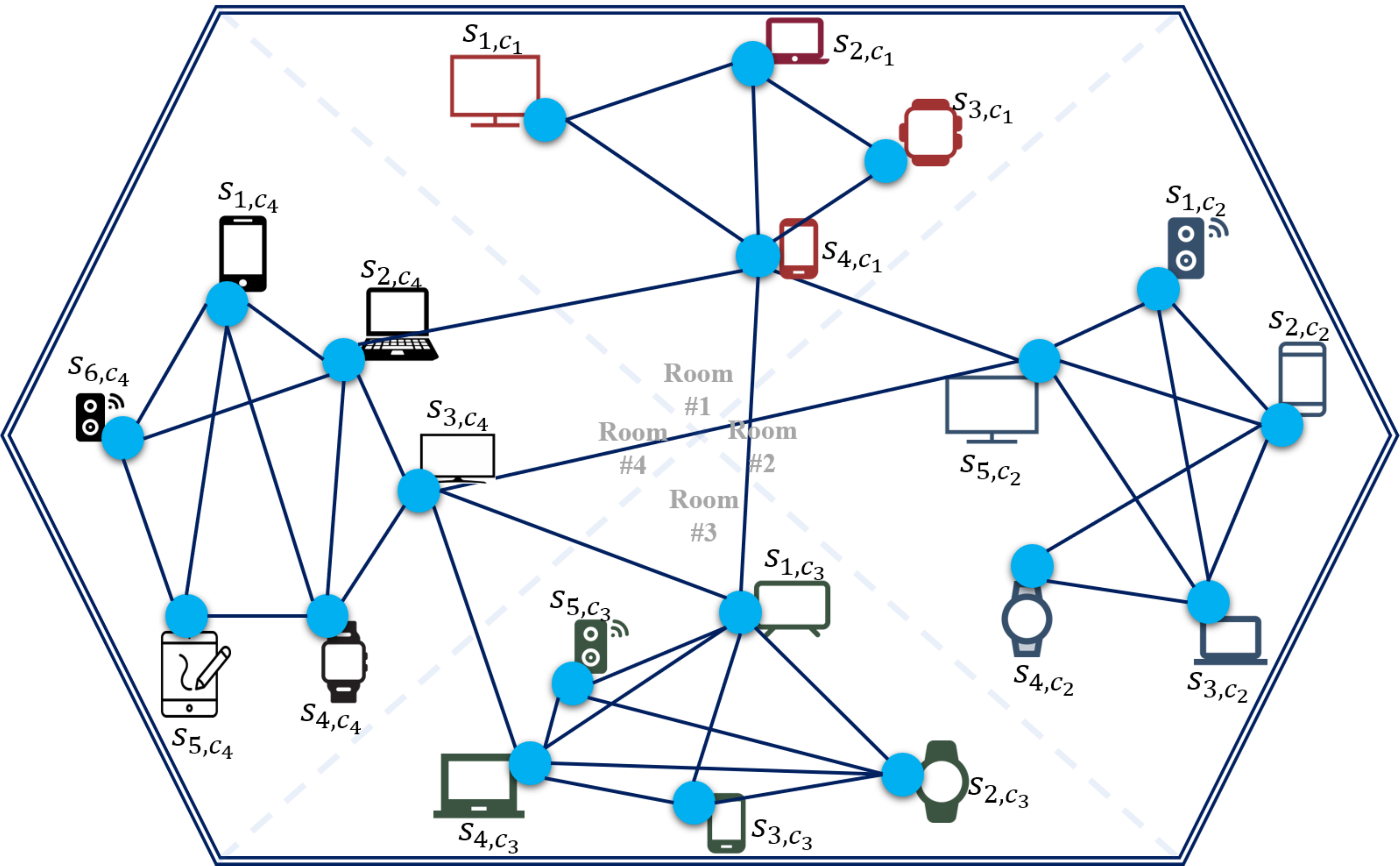}}
\caption{The graph structure used in the simulation. Each node represents a device with different performance capability and data/label heterogeneity. The distance between nodes in this figure does not indicate the actual distance between devices.}
\label{fig:graph_sim}
\end{figure}

\textcolor{black}{Furthermore, we presume a deterministic graph structure $\mathcal{G}$ in the experiments illustrated in Fig. \ref{fig:graph_sim}.} Note that the graph frequencies of $\mathcal{G}$ are the horizontal lines shown in Fig. \ref{fig:GF}. In particular, there are $20$ eigenvalues each corresponding to the graph frequencies where the first one with the value $0$ indicates the DC and the last one with the value $8.06$ corresponds to the highest graph frequency in $\mathcal{G}$.  Moreover, the power spectral density of the white noise is set to $n_0=-174$ dBm/Hz and the effective switched capacitance is fixed to $\varsigma=10^{-28}$ as suggested in \citep{mao2016dynamic}. To have a more realistic scenario, we sample heterogeneous values for the computing intensity, computation capacity, and the average transmit power from the uniform distribution $U$ as $\rho_{s_{i,c_j}}\sim U(1,3.5)$ cycles/sample, $f_{s_{i,c_j}}\sim U(1,3.5)$ GHz, and $P_{s_{i,c_j}}\sim U(0.5,1)$ W, respectively. Furthermore, we assume full participation where all devices share their models at each communication round. We also set $\alpha_{s_{i,c_j}}=3$ for the device-based number of iterations in all our experiments, except in the parameter optimization simulation. The whole parameter specifications are gathered in Table \ref{tab:sim_param} for convenience.
\begin{table}[]
\caption{Simulation parameters. }
\centering
\label{tab:sim_param}
\begin{tabular}{|c||c|}
\hline
Parameter & Value \\ \hline
$A_j \times B_j \times C_j$ & $10^3$ m$^3$ , $\forall c_j$ \\ \hline
$N$ & $4$ \\ \hline
$K_{c_j}$ & $\{K_{c_j}\sim Du(4,7)|K=20\}$ \\ \hline
$n_0$ & $-174$ dbm/Hz \\ \hline
$\rho_{s_{i,c_j}}$ & $U(1,5)\times 10^4$ cycles/sample \\ \hline
$\varsigma$ & $10^{-28}$  \\ \hline
$f_{s_{i,c_j}}$ & $U(1,3.5)$ GHz \\ \hline
$p^{\textrm{tran}}_{s_{i,c_j}}$ & $U(0.5,1)$ W \\ \hline
$\xi_{s_{i,c_j}}$ & $U(1,2)$ dB \\ \hline
$\beta$ & 20 MHz \\ \hline
\end{tabular}
\end{table}

\textbf{Heterogeneity indicator}: To have a quantitative representation of the system heterogeneity, we define
\begin{equation}\label{eq:H}
H= 1- \dfrac{1}{K}\sum_{j=1}^{N}\sum_{i=1}^{K_{c_j}}\dfrac{Min\{\widehat{\tau}_{s_{i,c_j}}\}}{\widehat{\tau}_{s_{i,c_j}}},
\end{equation}
where $\widehat{\tau}_{s_{i,c_j}}$ indicates the total computation and communication time given $\psi(\mathcal{D}_{s_{i,c_j}};q_{s_{i,c_j}})=1$ and $\chi(\varphi(\boldsymbol{g}_{s_{i,c_j}};z_{s_{i,c_j}}))=1$. The indicator in (\ref{eq:H}) suggests a high performance variability if $H \rightarrow 1$ and vice versa.

 In addition, we opt to assess the proposed framework for the image classification as an example of the computer vision application. In this regard, a deep model implemented in Tensorflow is used consisting of $32$ and $64$ Conv2D filters, respectively, each with the kernel size of $3$ and stride $2$ followed by a ReLu activation function and a maxpooling layer with $(2,2)$ pooling size. The last layer of this convolutional layer is then connected to the fully connected layers with the $128$ and $10$ neurons in each successive layer. It should be noted that the main goal in this work is not to achieve a state-of-the-art image classification accuracy. Here, we aim to compare the proposed approach in model personalization and the impact of graph filtering with different settings in our GFIoTL framework using the G-Fedfilt aggregation rule. Indeed, a more complex deep model can achieve higher accuracy, however, the relative results provided in this paper can still apply using a different deep model.
\subsection{Datasets}
We consider MNIST\footnote{http://yann.lecun.com/exdb/mnist/} and its extension, EMNIST datasets, that are commonly used for the evaluation of image classification tasks. \textcolor{black}{We leverage such generic datasets because they are standard and recognizable datasets for ML model evaluation. MNIST gives a moderate and typical complexity of IoT applications \citep{zhao2020privacy}. Furthermore, it is straightforward to create different levels of label heterogeneity, as a form of statistical heterogeneity to evaluate the framework. It is also noteworthy to mention that the MNIST dataset has been used extensively for testing FL frameworks and IoT systems \citep{xu2020lightweight,mills2019communication}.} MNIST consists of $60,000$ training and $10,000$ testing examples of $28 \times 28$ digit images with $0$ to $9$ as the labels. For the MNIST dataset, we take the same step as \citep{mcmahan2017communication} to create different label-heterogeneous Non-i.i.d./i.i.d. datasets. In particular, three datasets are created and  indicated respectively as MNIST$_2$, MNIST$_4$, and MNIST$_{10}$, where for MNIST$_2$ there only exist two classes per device, MNIST$_4$ four classes per device, and for MNIST$_{10}$ all classes exists at each device $s_{i,c_j} \in \mathcal{K}_{c_j}$. \textcolor{black}{Such label heterogeneity is quite common in smart home application scenarios. For instance, in the case of HAR, one client might avoid activities such as exercising and running. Consequently, the datasets collected by devices will only contain partial labels.} We choose $D_{s_{i,c_j}}=450$, $\forall s_{i,c_j}$, for the training dataset. To evaluate the personalization of the models over their local dataset as well as their generalization, we create two types of test sets. The first set indicated as ``local test set'' is created with the same label distribution as the training set with $100$ data samples; and the second one, ``global test set'', is created with all the labels to create $100$ data samples. Such a test set is important since, in personalized FL, it is often straightforward to notice a decrease in accuracy when data from other distributions is tested on a personalized model. Thus, a global test set could potentially indicate how much the generalization of the model is lost because of personalization.  

 The EMNIST dataset is by default, a heterogeneous set with data heterogeneity. \textcolor{black}{The dataset is similar to the situation in HAR applications where each client has its own way of performing an activity distinctive to others.} Here, we only considered the digit images for the training where each device has $D_{s_{i,c_j}}=450$ training samples. This dataset has a data heterogeneity (as oppose to label heterogeneity) since the authors have a different handwriting. In the case of the test set, we follow the same step as for the MNIST dataset where there are two local and global test sets.
 Throughout the simulations, we also average the model accuracy of edge devices to calculate a single accuracy. Furthermore, the following experiments and their results are the averages of $5$ times of simulation run.
\subsection{Performance Indices}
To assess the credibility of the proposed GFIoTL framework, we define various performance indices in terms of computation, communication, and \textcolor{black}{models' classification capabilities}. For better presentation, the time step $t$ is dropped in the formulations meaning that the parameters are prone to change at each communication round $t$. Hence, the metrics are expressed as follows:

\textbf{\textcolor{black}{Classification metrics:}}
 \textcolor{black}{We use four classification performance metrics to evaluate the behavior of the trained models. To do so, we first define an $n_c \times n_c$ confusion matrix $CM_{s_{i,c_j}}$ for each model in device $s_{i,c_j}$ where $n_c$ is the total number of classes. The elements of the confusion matrix are filled according to the true and predicted labels. Overall, $K$ confusion matrices are created due to having $K$ devices. We then apply a summation to construct a confusion matrix $CM^{\textrm{total}}=\sum_{j=1}^{N}\sum_{i=1}^{K_{c_j}}CM_{s_{i,c_j}}$ in which, all the test samples in the framework are encompassed. Additionally, $CM^{\textrm{total}}$ can be interpreted through metrics such as accuracy, precision, recall, and F1-score. We define each metric, respectively, as 
\begin{equation}
I_1= \dfrac{1}{n^{\textrm{total}}}\sum_{i=1}^{n_c} TP_i,
\end{equation}
\begin{equation}
I_2= \dfrac{1}{n_{c}}\sum_{i=1}^{n_c}\dfrac{TP_i}{TP_i+FP_i},
\end{equation}
\begin{equation}
I_3 = \dfrac{1}{n_{c}}\sum_{i=1}^{n_c}\dfrac{TP_i}{TP_i+FN_i},
\end{equation}
\begin{equation}
I_4= \dfrac{2I_2 \times I_3}{I_2+I_3},
\end{equation}
where $TP_i$, $FP_i$, and $FN_i$ represent true positives, false positives, and false negatives in $CM^{\textrm{total}}$ associated with $i^{th}$ class. Moreover, $n^{\textrm{total}}$ is the total number of samples in the test set used to evaluate the models.}

\textbf{Computation Cost:}
This metric determines how much computation is spent on the local CIoT devices as a whole. The computation cost can be evaluated as the total sum of Floating-Point Operations (FLOPs) executed on the edge side expressed as 
\begin{equation}
I_5 = \sum_{t=1}^{R} \sum_{j=1}^{N}\sum_{i=1}^{K_{c_j}}\alpha_{s_{i,c_j}} \phi_{s_{i,c_j}} \psi(\mathcal{D}_{s_{i,c_j}};q_{s_{i,c_j}}),
\end{equation}
where $\phi_{s_{i,c_j}}$ denotes the FLOPs per one input data sample for device $s_{i,c_j}\in \mathcal{K}_{c_j}$.


\textbf{Communication Latency:}
In FL, the slowest device in the framework always drags the whole training procedure causing the deadline $T$ to become a large value.  In this regard, the total latency of an FL framework is given by
\begin{equation}
I_6 = \sum_{t=1}^{R}T .
\end{equation} 
\textbf{Communication Desynchronization:}
 Since CIoT devices have different computational capabilities and might experience different fading channels, the gradient updates will not reach the server at the same time. In addition, different distances between edge devices and the server along with poor connections exacerbate this phenomenon and prevent a synchronized aggregation. This causes the edge server to be active for a longer period of time in order to collect all the gradients. Therefore, it is desirable to minimize the total desynchronization time at the server side for $R$ rounds defined as:
\begin{equation}
I_7 = \sum_{t=1}^{R} \left(T- \min \{ \tau_{s_{i,c_j}}\}\right).
\end{equation}
\subsection{Results and Discussions}
In this subsection, the performance of the proposed GFIoTL framework under various conditions is investigated. We first assess the embedded G-Fedfilt aggregation rule as the main backbone of the proposed structure. G-Fedfilt is compared with FedAvg and evaluated with different $h_s$ using statistically heterogeneous datasets. Afterward, we evaluate the impact of parameter optimization and the subsequent solutions acquired from problem $\mathcal{P}$ in terms of the aforementioned performance indices. 

\begin{figure}[!tb]
\centerline{\includegraphics[scale=.6]{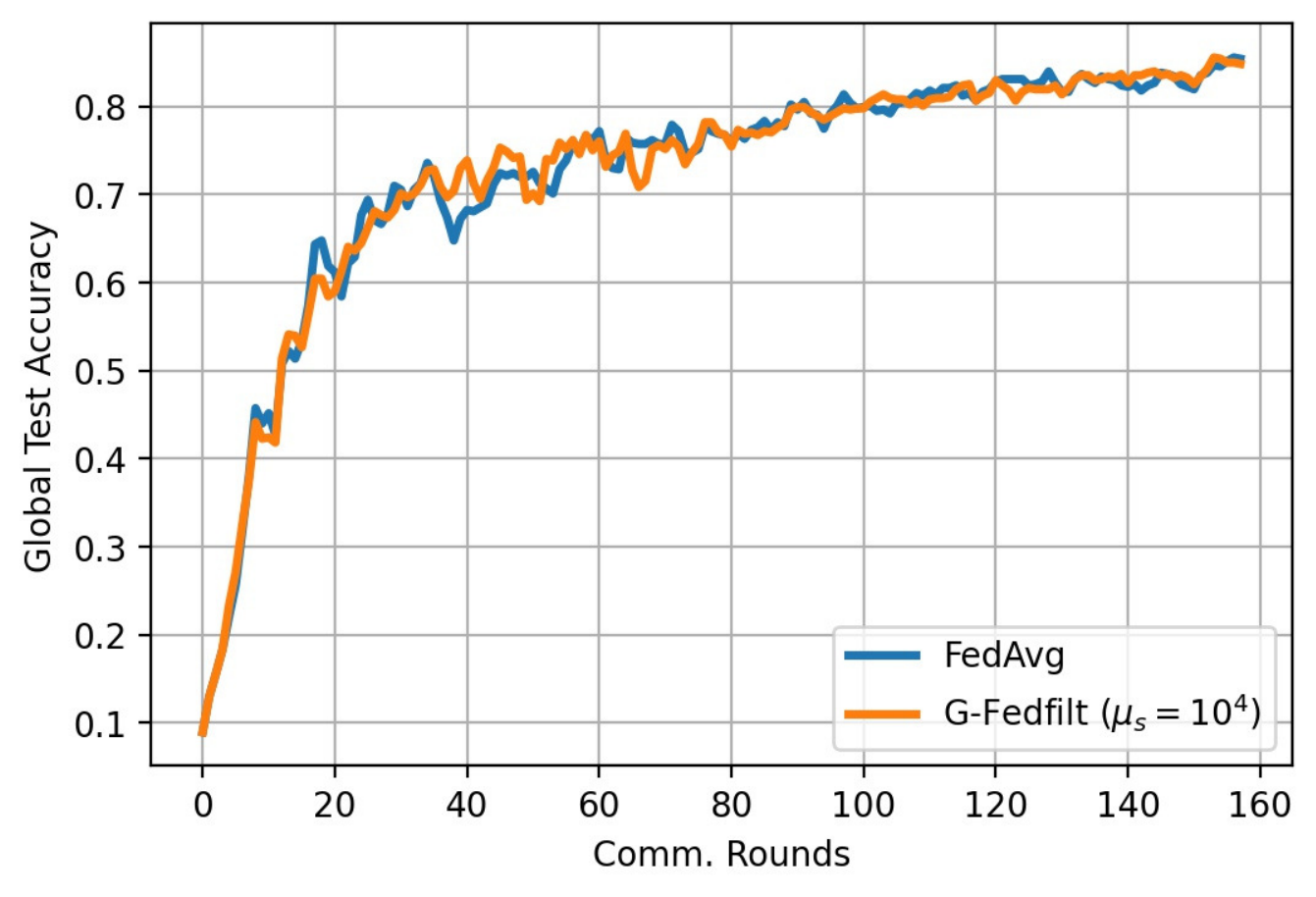}}
\caption{Classification accuracy of the proposed G-Fedfilt and FedAvg algorithms trained on EMNIST dataset where the graph filter is designed to pass only the DC components of the gradient update matrix.}
\label{fig:AvgvsG-FedFilt}
\end{figure}

\textbf{G-Fedfilt as FedAvg:}
Here, we investigate the relationship between G-Fedfilt and FedAvg and how it incorporates FedAvg aggregation rule. In this case, we choose $h_s(\lambda;\mu_s=10^{4})$ in order to have a complete domain-agnostic aggregation. Fig. \ref{fig:AvgvsG-FedFilt} shows the results of FedAvg and G-Fedfilt on the EMNIST global test set. As observed, both G-Fedfilt and FedAvg achieve almost the same classification accuracy during each round of communication. Of course, the learning curves are not exactly overlapping and this is primarily due to the Tensorflow implementation and the usage of built-in functions for the training rather than the theoretical aspect of the proposed method.

\begin{figure*}[t!]
\centering
\subfloat[\label{fig:rr2}]{\includegraphics[width=0.45\textwidth]{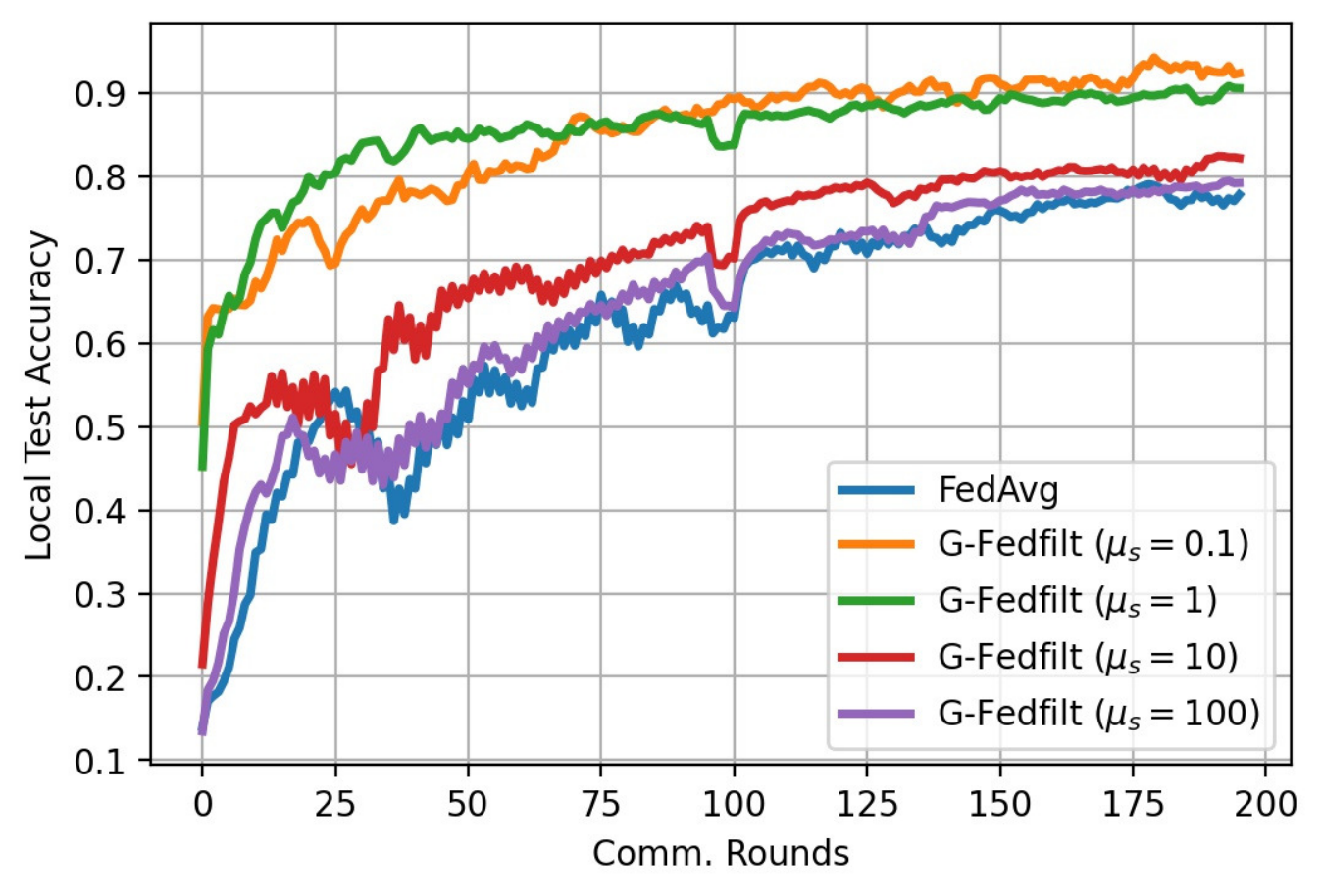}}\hfill
\subfloat[\label{fig:rr3}] {\includegraphics[width=0.45\textwidth]{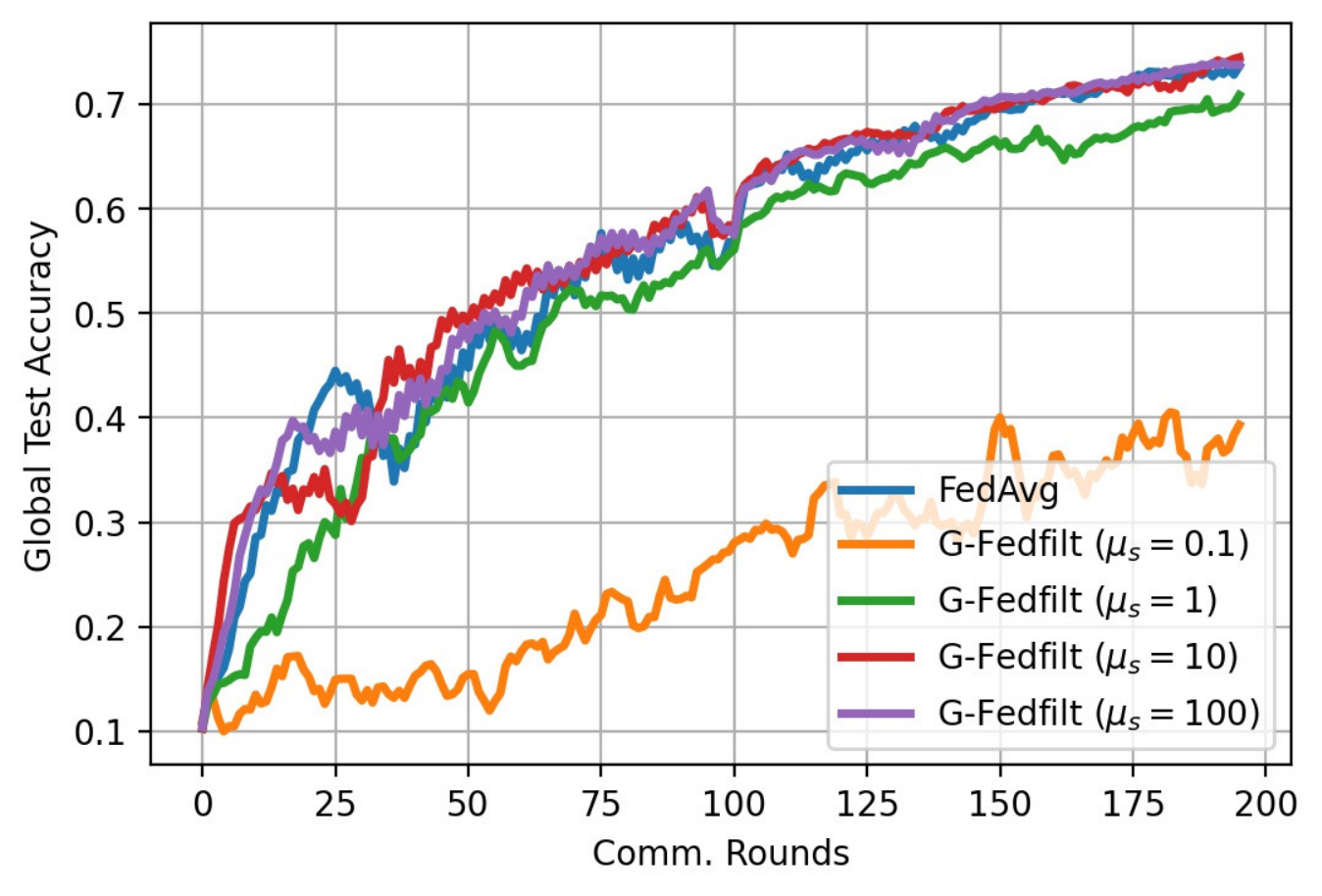}}\hfill
\subfloat[\label{fig:rr4}]{\includegraphics[width=0.45\textwidth]{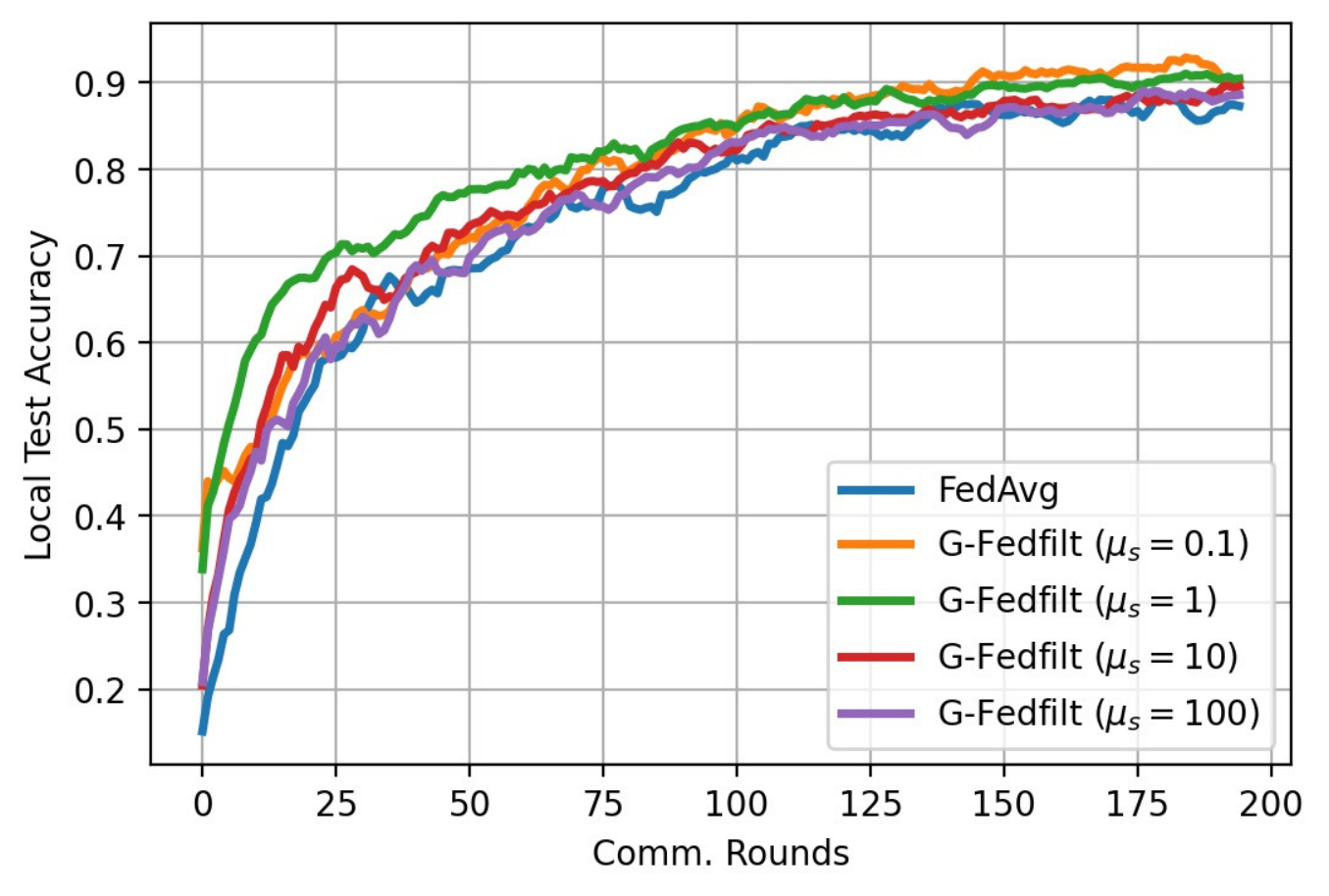}}\hfill
\subfloat[\label{fig:rr5}]{\includegraphics[width=0.45\textwidth]{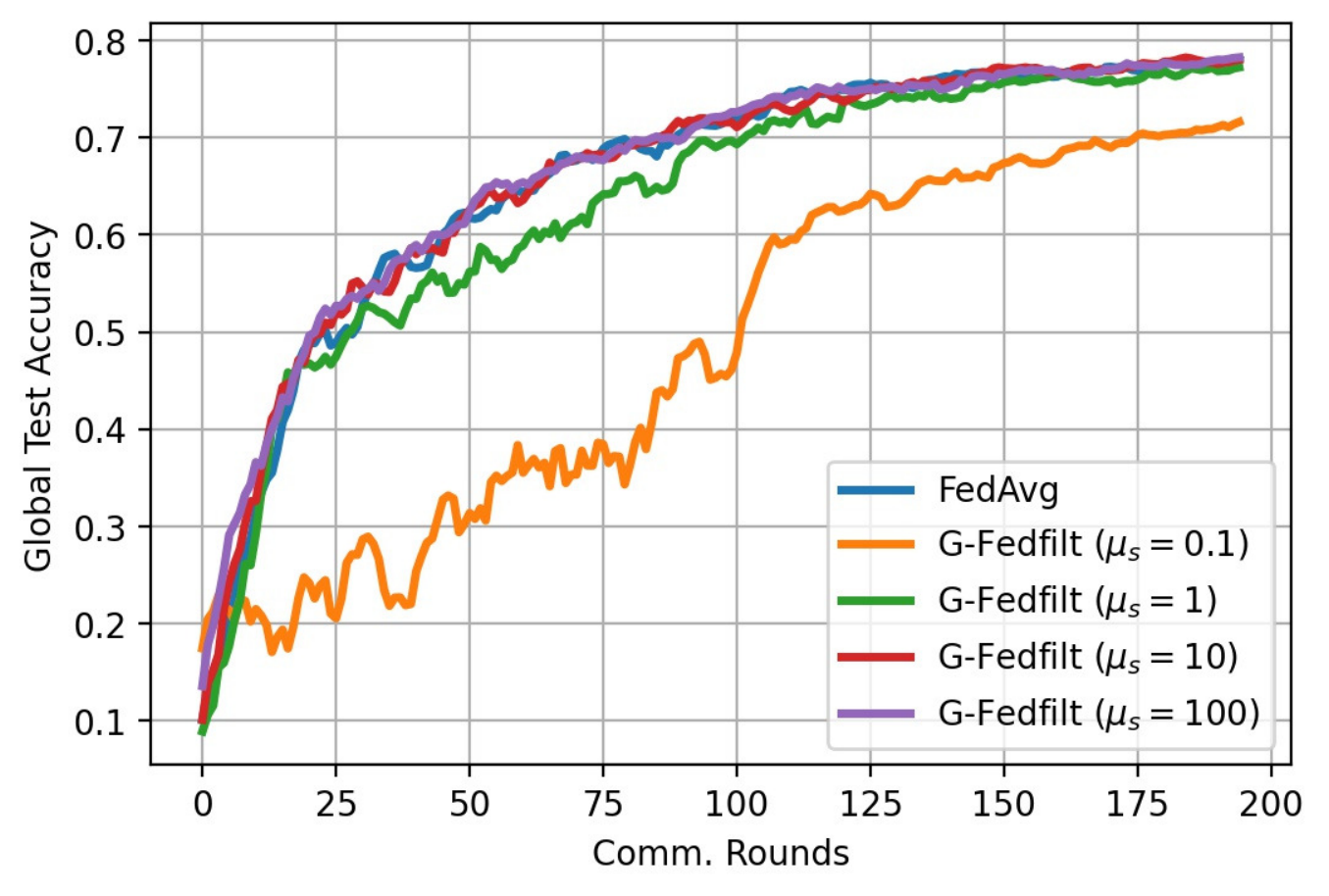}}
\caption{Comparing the performance of the proposed G-Fedfilt aggregation rule with FedAvg using different values of $\mu_s$. \textcolor{black}{The models are trained on statistically heterogeneous MNIST$_2$ (a,b) and MNIST$_4$ (c,d) datasets} and are evaluated on both the global (right column) and the local (left column) test sets.} \label{fig:R2}
\end{figure*}

\textbf{Convergence Behavior:}
To understand the impact of different graph filtering on the convergence curves, we run the simulation on MNIST$_2$ and MNIST$_4$ datasets by adjusting various values for $\mu_s$. Fig. \ref{fig:R2} shows the convergence curve of different settings tested on both the local and the global test sets for $200$ communication rounds. In Fig. \ref{fig:R2}a, it is inferred that decreasing the value of $\mu_{s}$ results in more personalization of the models on the local test set at the edge. This is mainly because by adjusting a low value for $\mu_{s}$, higher graph frequencies remain involved. On the other hand, from Fig. \ref{fig:R2}b, it can be seen that the generalization of the models on the global test set often reduces with lower $\mu_s$; however, for some parameters such as $\mu_{s}=1$ or $\mu_{s}=10$, the generalization keeps improving close to FedAvg over the communication rounds. In other words, G-fedfilt is capable of retaining the personalization over edge devices while keeping a decent level of generalization in the models. The same argument applies for the results tested on MNIST$_4$ shown in Fig. \ref{fig:R2}c,d; however, since there is less statistical heterogeneity involved, we do not see large differences between convergence curves evaluated on the global and local test sets as much as Fig. \ref{fig:R2}a,b.

\begin{table*}[]
\centering
\caption{The quantitative results of FedAvg and G-Fedfilt using different parameter settings evaluated on the local test sets and after $200$ communication rounds. The values are indicated as the mean $\pm$ std of the accuracy of $K=20$ devices.}
\label{tab:local_test}
\begin{tabular}{cc|cccc}
\hline
\multirow{2}{*}{Algorithms} & \multirow{2}{*}{Parameters} & \multicolumn{4}{c}{Datasets} \\ \cline{3-6} 
 &  & \multicolumn{1}{c|}{MNIST$_{10}$} & \multicolumn{1}{c|}{MNIST$_4$} & \multicolumn{1}{c|}{MNIST$_2$} & EMNIST \\ \hline
FedAvg & ---- & \multicolumn{1}{c|}{$86.12 \pm 4.84$} & \multicolumn{1}{c|}{$88.37 \pm 4.95$} & \multicolumn{1}{c|}{$79.50 \pm 10.32$} & $95.99\pm 6.79$ \\ \hline
\multirow{4}{*}{G-Fedfilt} & $\mu_{s}=0.1$ & \multicolumn{1}{c|}{$79.74\pm 7.91$} & \multicolumn{1}{c|}{$92.74 \pm 4.34$} & \multicolumn{1}{c|}{$95.62\pm 9.52$} & $96.58\pm 6.44$ \\
 & $\mu_{s}=1$ & \multicolumn{1}{c|}{$84.37\pm 5.55$} & \multicolumn{1}{c|}{$91.12 \pm 4.20$} & \multicolumn{1}{c|}{$92.37 \pm 8.14$} & $97.41 \pm 6.48$ \\
 & $\mu_{s}=10$ & \multicolumn{1}{c|}{$85.50 \pm 6.25$} & \multicolumn{1}{c|}{$89.87 \pm 4.43$} & \multicolumn{1}{c|}{$83.49\pm 9.54$} & $96.16 \pm 6.84$ \\
 & $\mu_{s}=100$ & \multicolumn{1}{c|}{$85.75 \pm 5.37$} & \multicolumn{1}{c|}{$89.35 \pm 4.82$} & \multicolumn{1}{c|}{$80.49 \pm 10.64$} & $95.59 \pm 6.84$ \\ \hline
\end{tabular}
\end{table*}
\begin{table*}[]
\centering
\caption{The quantitative results of FedAvg and G-Fedfilt using different parameter settings evaluated on the global test set and after $200$ communication rounds. The values are indicated as the mean $\pm$ std of the accuracy of $K=20$ devices.}
\label{tab:global_test}
\begin{tabular}{cc|cccc}
\hline
\multirow{2}{*}{Algorithms} & \multirow{2}{*}{Parameters} & \multicolumn{4}{c}{Datasets} \\ \cline{3-6} 
 &  & \multicolumn{1}{c|}{MNIST$_{10}$} & \multicolumn{1}{c|}{MNIST$_4$} & \multicolumn{1}{c|}{MNIST$_2$} & EMNIST \\ \hline
FedAvg & ---- & \multicolumn{1}{l|}{$85.05 \pm 0.00$} & \multicolumn{1}{c|}{$77.27 \pm 0.00$} & \multicolumn{1}{c|}{$76.04 \pm 0.00$} & $87.21 \pm 0.00$ \\ \hline
\multirow{4}{*}{G-Fedfilt} & $\mu_{s}=0.1$ & \multicolumn{1}{c|}{$80.55 \pm 6.56$} & \multicolumn{1}{c|}{$72.33 \pm 4.78$} & \multicolumn{1}{c|}{$49.03 \pm 7.45$} & $80.59\pm 2.02$ \\
 & $\mu_{s}=1$ & \multicolumn{1}{c|}{$84.49 \pm 1.25$} & \multicolumn{1}{c|}{$77.27 \pm 1.73$} & \multicolumn{1}{c|}{$73.07 \pm 5.38$} & $87.38\pm 0.52$ \\
 & $\mu_{s}=10$ & \multicolumn{1}{c|}{$86.02\pm 0.18$} & \multicolumn{1}{c|}{$77.75 \pm 0.34$} & \multicolumn{1}{c|}{$78.45 \pm 0.24$} & $88.09\pm 0.15$ \\
 & $\mu_{s}=100$ & \multicolumn{1}{c|}{$85.57 \pm .15$} & \multicolumn{1}{c|}{$78.04 \pm 0.07$} & \multicolumn{1}{c|}{$74.95 \pm 0.08$} & $88.14\pm 0.01$ \\ \hline
\end{tabular}
\end{table*}

\textbf{Effect of $\mu_s$ Under Statistical Heterogeneity:}
To have a full investigation on the tunable parameter $\mu_s$ and its behavior under data/label heterogeneity, we assess the performance of the proposed G-Fedfilt on i.i.d. dataset such as MNIST$_{10}$, and Non-i.i.d. datasets including MNIST$_{2}$, MNIST$_{4}$, and EMNIST. Table \ref{tab:local_test} shows the classification accuracy (mean $\pm$ standard deviation) of the proposed G-Fedfilt algorithm and FedAvg for $K=20$ devices and after $200$ communication rounds evaluated on the local test set. We observe that when the label and data heterogeneity is involved, G-Fedfilt achieves better accuracy than Fedavg, from $1.42\%$ for the EMNIST up to $6.12\%$ for the MNIST$_2$ datasets; while for the MNIST$_{10}$ where there is no heterogeneity, it does not perform superior to FedAvg on the local test set. It is also seen that increasing $\mu_s$ would reduce the accuracy since by doing so, higher frequencies are filtered. On the other hand, when looking at Table \ref{tab:global_test}, which shows the resultant accuracy on the global test set, we see that imposing less personalization (by increasing $\mu_s$) often eventuates in a better classification accuracy. This argument fairly applies to all the datasets. \textcolor{black}{In addition, Table \ref{tab:F1} shows the classification performance based on precision, recall, and F1-score. Here, the models are trained on the heterogeneous MNIST$_2$ training dataset and tested on its local and global test sets. As seen, assigning a lower value to $\mu_s$ shows better performance on the local test set. On the other hand, the reduction in $\mu_s$ decreases the performance on the global test set in all three metrics. Based on the results in Tables \ref{tab:local_test}, \ref{tab:global_test}, and \ref{tab:F1}, it is rather concrete to say that there is a trade-off between model personalization and generalization in the G-Fedfilt aggregation rule that can be adjusted by the parameter $\mu_s$.}

\begin{figure}[!t]
\centerline{\includegraphics[scale=.6]{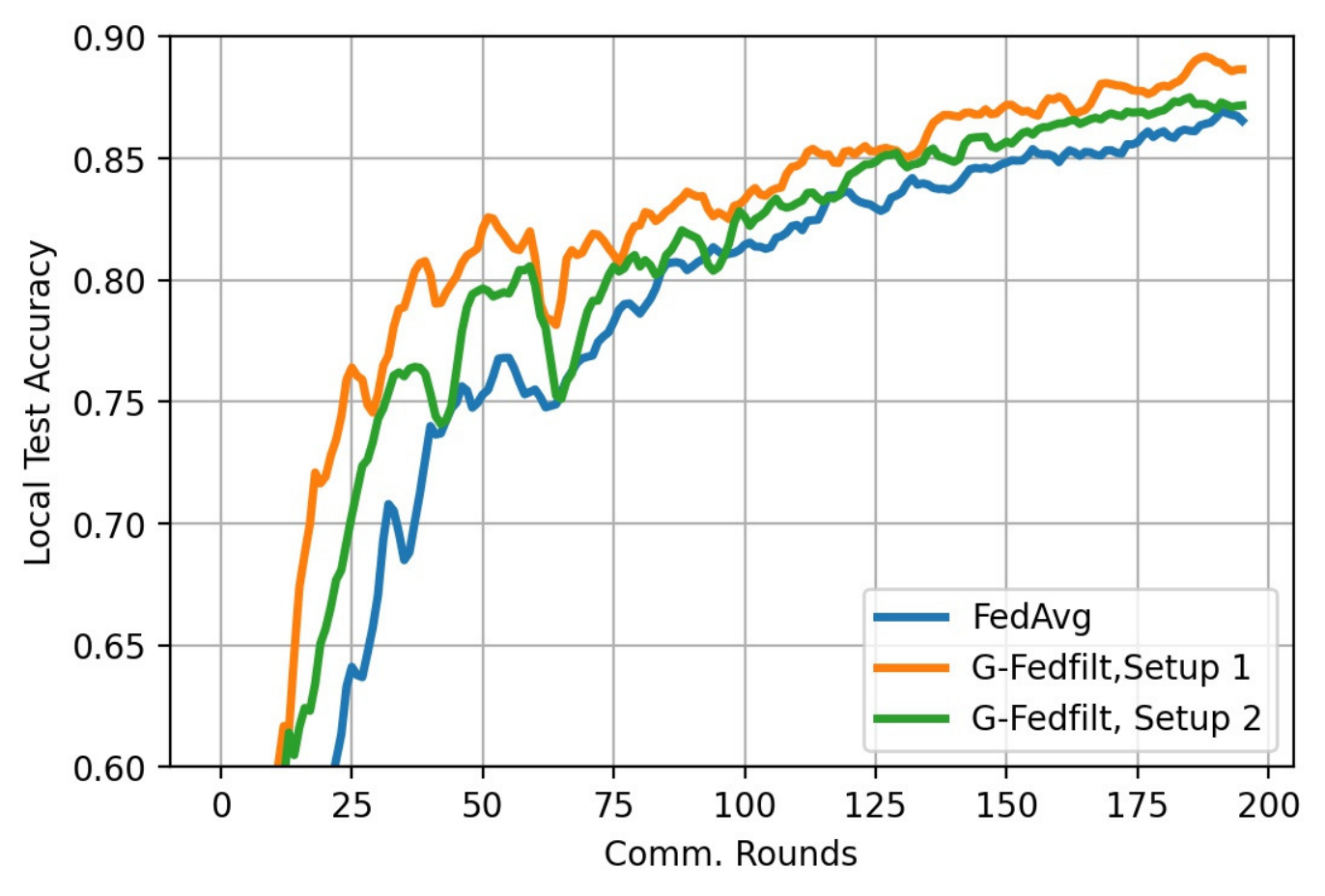}}
\caption{Comparison of the proposed G-Fedfilt and FedAvg convergence curve in two setups on MNIST$_4$ dataset. In Setup 1, edge devices in a certain cluster have similar data distribution while, in Setup 2, their data distributions are randomly chosen. }
\label{fig:twosetups}
\end{figure}

\textbf{Effect of Graph Connectivity:}
Here, we explore the role of devices' connectivity on the model personalization via a graph. We distribute the MNIST dataset such that each device $s_{i,c_j}$ in cluster $c_j \in \mathcal{N}$ has similar label distribution as other devices in $c_j$ while, exhibits a different distribution with devices of other clusters. More specifically, the device $s_{i_1,c_j}$ has a dataset consisting of $4$ labels, $3$ of which are labeled the same as the device $s_{i_2,c_j}$. Therefore, the devices connected in a particular cluster tend to have the same data distribution. We call this ``Setup 1'' to further recall that in the simulation. Note that this behavior of the same distribution in a cluster is not far from a real-world scenario. \textcolor{black}{This is because the devices in a certain cluster (or a smart room/home) are often owned by a particular client which makes the data gathered by edge devices have a similar distribution.} To compare this scenario with a situation where devices in a cluster do not have a similar distribution, we created ``Setup 2'' where we randomly selected $4$ labels out of $10$ for each device without the consideration of their positions in the graph. These setups are then used for training. We choose FedAvg and G-Fedfilt with $h_s(\lambda;\mu_{s}=10)$ for the aggregation in order to have a fair comparison irrespective to the graph filter. Fig. \ref{fig:twosetups} illustrates the convergence curve of such simulation evaluated on the local test set. For better inspection, the y-axis of this figure starts from $0.6$. As seen, G-Fedfilt in Setup 1 is consistently better compared to its performance in Setup 2. Furthermore, when using G-Fedfilt, both setups can achieve better accuracy than FedAvg. Note that FedAvg performs the same in both setups since it does not involve the connections of devices. Hence, it is concluded that the graph connectivity in the proposed GFIoTL framework plays an important role in the model personalization.  
\begin{table*}[]
\centering
\caption{\textcolor{black}{Performance evaluation of the proposed G-Fedfilt and FedAvg algorithms in terms of precision, recall, and F1-score. The algorithms are evaluated on the MNIST$_2$ dataset after $R=200$ communication rounds. The metrics are indicated in \%.}}
\label{tab:F1}
\begin{tabular}{cc|ccc|ccc}
\hline
\multirow{2}{*}{Algorithm} & \multirow{2}{*}{Parameter} & \multicolumn{3}{c|}{\begin{tabular}[c]{@{}c@{}} Local test set\end{tabular}} & \multicolumn{3}{c}{\begin{tabular}[c]{@{}c@{}}Global test set\end{tabular}} \\ \cline{3-8} 
                           &                           & \multicolumn{1}{c|}{Precision}       & \multicolumn{1}{c|}{Recall}      & F1-score      & \multicolumn{1}{c|}{Presicion}       & \multicolumn{1}{c|}{Recall}      & F1-score      \\ \hline
FedAvg                     & \_\_\_                    & \multicolumn{1}{c|}{71.99}           & \multicolumn{1}{c|}{77.70}       & 74.74         & \multicolumn{1}{c|}{72.07}           & \multicolumn{1}{c|}{75.03}       & 73.52         \\ \hline
\multirow{4}{*}{G-Fedfilt} & $\mu_s=0.1$                & \multicolumn{1}{c|}{92.85}           & \multicolumn{1}{c|}{93.72}       & 93.29         & \multicolumn{1}{c|}{34.93}           & \multicolumn{1}{c|}{43.17}       & 38.61         \\
                           & $\mu_s=1$                  & \multicolumn{1}{c|}{87.14}           & \multicolumn{1}{c|}{89.68}       & 88.39         & \multicolumn{1}{c|}{66.53}           & \multicolumn{1}{c|}{73.20}       & 69.70         \\
                           & $\mu_s=10$                 & \multicolumn{1}{c|}{74.58}           & \multicolumn{1}{c|}{81.78}       & 78.01         & \multicolumn{1}{c|}{71.73}           & \multicolumn{1}{c|}{74.41}       & 73.04         \\
                           & $\mu_s=100$                & \multicolumn{1}{c|}{73.91}           & \multicolumn{1}{c|}{79.36}       & 76.54         & \multicolumn{1}{c|}{72.51}           & \multicolumn{1}{c|}{75.49}       & 73.96  
\\ \hline      
\end{tabular}
\end{table*}

\begin{table*}[]
\centering
\caption{Performance evaluation of the proposed G-Fedfilt in terms of classification accuracy, communication desybchronization time, and latency before and after the parameter optimization is tested on MNIST$_{10}$ global test set under various system heterogeneity settings.}
\label{tab:sysHetResults}
\resizebox{\columnwidth}{!}{%
\begin{tabular}{c|c|ccc|ccc|c}
\hline
\multirow{2}{*}{Heterogeneity} & \multirow{2}{*}{Algorithm} & \multicolumn{3}{c|}{After 200 comm. rounds} & \multicolumn{3}{c|}{After 400 Comm. Rounds} & \multirow{2}{*}{$\Delta$  Acc.(\%)} \\ \cline{3-8}
 &  & Acc. (\%) & Comm. Desync. (s) & Latency (s) & Acc. (\%) &  Comm.  Desync. (s) &  Latency (s) &  \\ \hline
\multirow{2}{*}{$H = 0.31$} & G-Fedfilt & 86.66 & 167.30 & 349.18 & 88.01 & 319.8 & 680.34 & $+1.35$ \\
 & G-Fedfilt$_{Opt}$ & 84.52 & 0.62 & 181.87 & 87.65 & 1.11 & 356.46 & $+3.13$ \\ \hline
\multirow{2}{*}{$H = 0.54$} & G-Fedfilt & 86.66 & 453.00 & 661.78 & 88.01 & 921.51 & 1332.92 & $+1.35$ \\
 & G-Fedfilt$_{Opt}$ & 82.24 & 12.68 & 220.04 & 86.25 & 25.82 & 443.17 & $+4.01$ \\ \hline
\multirow{2}{*}{$H =0.65$} & G-Fedfilt & 86.66 & 861.59 & 1009.50 & 88.01 & 1706.69 & 2004.61 & $+1.35$ \\
 & G-Fedfilt$_{Opt}$ & 74.41 & 55.09 & 201.90 & 83.05 & 108.23 & 399.04 & $+8.64$ \\ \hline
\end{tabular}%
}
\end{table*}

\begin{figure}[!t]
\centerline{\includegraphics[scale=.4]{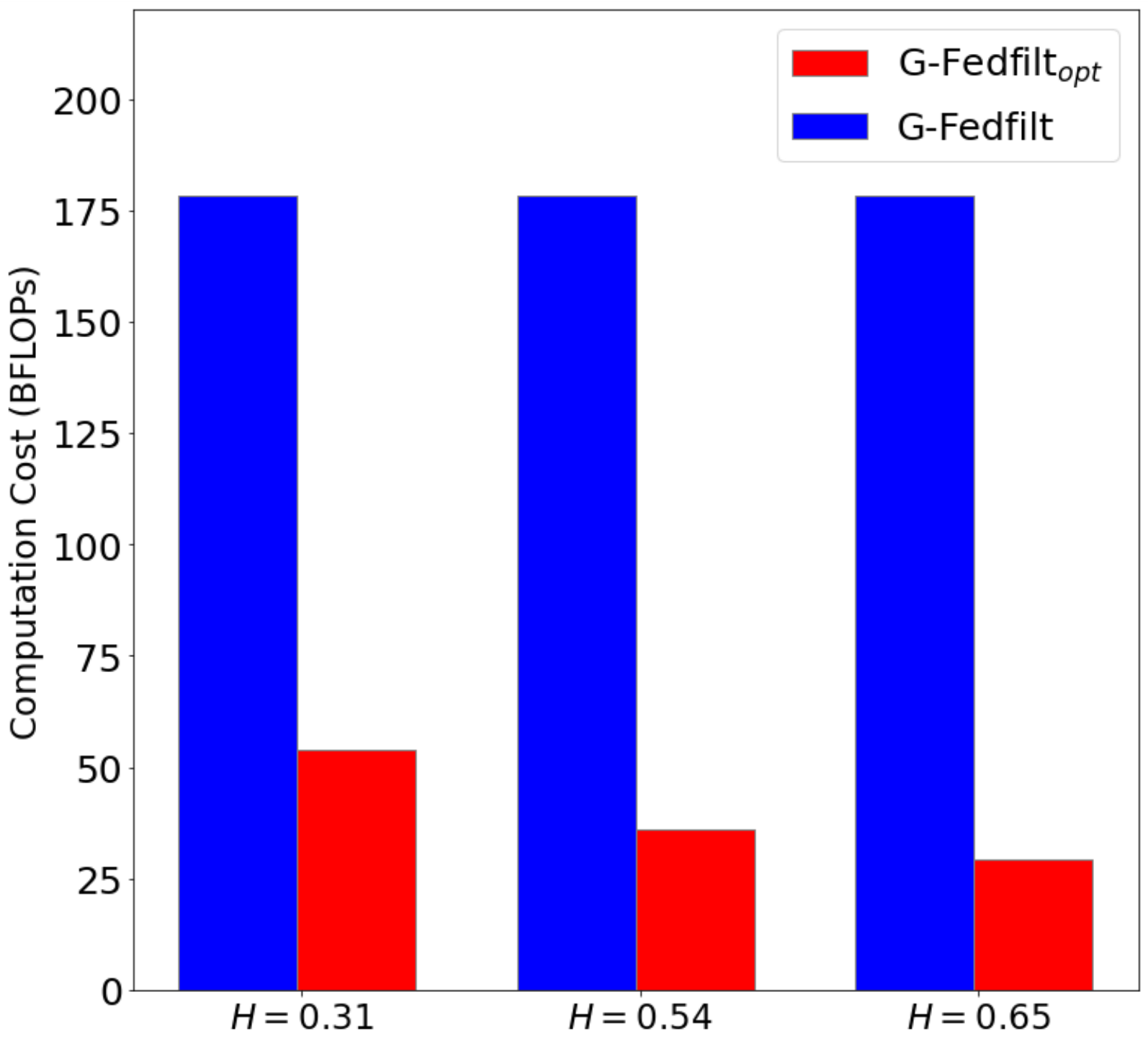}}
\caption{\textcolor{black}{The computation cost of CIoT devices under different system heterogeneity in terms of Billion FLOPs (BFLOBs) after $400$ communication rounds. }}
\label{fig:bargraph}
\end{figure}

\textbf{Communication Desynchronization Alleviation:}
In this simulation, we investigate the effect of dynamic values for $\psi(\mathcal{D}_{s_{i,c_j}};q_{s_{i,c_j}})$ and $\chi(\varphi(\boldsymbol{g}_{s_{i,c_j}};z_{s_{i,c_j}}))$ on the communication desynchronization and latency when the control parameters $z_{s_{i,c_j}}$, $q_{s_{i,c_j}}$, and $\alpha_{s_{i,c_j}}$ are optimized based on solving problem $\mathcal{P}$. We set $\mu_1 = \mu_2=0.4$ and $\mu_3=0.2$ for solving the optimization problem in the simulation where $\mu_1+\mu_2+\mu_3=1$. Note that the reason for selecting a smaller $\mu_3$ lies behind the fact that it is  experimentally shown in \citep{aji2017sparse}, that almost $90\%$ of the gradients to be sent to the server can be ignored with nearly no loss of classification accuracy. In addition, maximizing $\psi(\mathcal{D}_{s_{i,c_j}};q_{s_{i,c_j}})$ and $\alpha_{s_{i,c_j}}$ participate more to higher classification accuracy compared to that of $z_{s_{i,c_j}}$ which corresponds to decreasing the sparsification. \textcolor{black}{We further opt to select $b_{s_{i,c_j}}=\frac{\beta}{K}$}, $q_{s_{i,c_j}}^{\textrm{min}}=0.3$, $\alpha_{s_{i,c_j}}^{\textrm{min}}=1$, $\alpha_{s_{i,c_j}}^{\textrm{max}}=5$, and $z_{s_{i,c_j}}^{\textrm{min}}=0.1$, $\forall s_{i,c_j} \in \mathcal{K}_{c_j}$, for simplicity and to have a minimum convergence requirement for the training process. Thus, after the optimization, the number of data samples for training and the sparsified gradients required to be sent to the server is obtained for each device. We determine different heterogeneous settings, indicated by $H$, by producing random values for simulation parameters specified in Table \ref{tab:sim_param}. The performance of the proposed G-Fedfilt algorithm with $\mu_s=10$ is evaluated in Table \ref{tab:sysHetResults} before and after the framework is optimized for different $H$. Moreover, the models are trained on MNIST$_{10}$ dataset and evaluated on its global test set. After $200$ communication rounds, we can observe that there is a $99.63\%$ and $47.91\%$ reduction in the total communication desynchronization and latency when comparing G-Fedfilt$_{Opt}$ with G-Fedfilt for $H=0.31$; although, it costs $2.14\%$ accuracy reduction. It is also seen that when $H$ increases, the desynchronization time and the latency are affected less for G-Fedfilt$_{Opt}$ than that of G-Fedfilt; however, the accuracy decreases largely. One more important result that can be deduced in Table \ref{tab:sysHetResults} is that when executing the algorithm for $400$ communication rounds, the accuracy of G-Fedfilt$_{Opt}$ increases at a faster pace than that of G-Fedfilt. This phenomenon is due to the fact that at each round of communication some devices with less computational capabilities train their models with the subset of the true dataset, i.e., $\widehat{\mathcal{D}}_{s_{i,c_j}}\subseteq \mathcal{D}_{s_{i,c_j}}$. Thus, G-Fedfilt$_{Opt}$ eventually achieves comparable accuracy to G-Fedfilt in the long run. It is worthwhile to note that a larger number of communication rounds does not mean a longer latency. For instance, we can see in Table \ref{tab:sysHetResults} that G-Fedfilt$_{Opt}$ achieves $+1\%$ better accuracy than G-Fedfilt with the fairly similar latency, however, $200$ more communication rounds. \textcolor{black}{In addition, the bar graph in Fig. \ref{fig:bargraph} shows the computation cost of this simulation under different heterogeneous settings and after $400$ communication rounds. As observed, the cost of G-Fedfilt$_{opt}$ is significantly lower than that of G-Fedfilt which suggests that the optimized G-Fedfilt is also computationally efficient.}

While it appears that the parameter optimization in the proposed GFIoT scheme reduces the classification accuracy, from a practical perspective, \textcolor{black}{the cost of computation and communication} might impose more than that of the accuracy after it surpasses a certain value. This is also intensified when CIoT devices are involved since they tend to relocate or pull out of the framework by the owners in the smart building. Another point to highlight here is the fact that due to the same relocation, device pull-out situation, or communication failure the graph representation of the network becomes more dynamic than deterministic as time elapses. Hence, the training procedure must be carried out as quickly as possible in order to approximate the dynamic graph as fixed in a limited time interval. Although lessening the latency with the proposed optimization problem might be one avenue, another fundamental approach is to design a graph-based FL for CIoT devices that is robust to such dynamics which we will leave to be addressed in our upcoming works. 
\section{Conclusion} \label{Se:conclusion}
In this paper, we introduced the Graph Federated Internet of Things Learning (GFIoTL) framework for collaborative training of CIoT devices in a smart building. To alleviate the effect of statistical heterogeneity, we developed a GSP-inspired aggregation rule based on graph filtering (G-Fedfilt) that incorporates the underlying connectivity of smart CIoT devices. This concept could potentially open up a wide range of filter designs for better model optimization and tunable personalization. Furthermore, the proposed GFIoTL framework is equipped with a communication-efficient parameter optimization scheme in order to lessen the impact of system heterogeneity in the framework. According to the simulation results, when tuned appropriately, G-Fedfilt was capable of achieving model personalization while attaining a decent amount of generalization in contrast to the conventional Federated Averaging (FedAvg). In addition, when parameter optimization was applied, \textcolor{black}{the computation cost, communication desynchronization, and latency} were reduced dramatically at the cost of a small reduction in classification accuracy.

\section*{Acknowledgment}This work was partially supported by the Natural Sciences and Engineering Research Council (NSERC) of Canada.

\section{Appendix}
  \subsection{The Mechanics Behind G-Fedfilt Algorithm}
  \label{FirstAppendix}
	To better understand the proposed G-Fedfilt algorithm, a simplified example is presented as follows. Consider an undirected light graph with $N =3$ and an adjacency matrix $\boldsymbol{A}_{3 \times 3}$. Each node/device $i$ has a model with a vector of two variables $\boldsymbol{g_i}=[g_{i1},g_{i2}]$. The Laplacian matrix of $\boldsymbol{A}$ is derived as
\begin{equation}
\boldsymbol{L}=\boldsymbol{D}-\boldsymbol{A}=\begin{bmatrix}
d_1-a_{11} & -a_{12} & -a_{13} \\ 
-a_{21} & d_2-a_{22} & -a_{23} \\ 
-a_{31} & -a_{32} &  d_3-a_{33} \\ 
\end{bmatrix},
\end{equation}
where $d_i=\sum_{j=1}^{3}a_{ij}$ is the degree of each node. We then decompose $\boldsymbol{L}_{3 \times 3}$ as $\boldsymbol{L}=\boldsymbol{V}\boldsymbol{\Lambda}\boldsymbol{V}^T$ and assume that it has ordered eigenvalues as $\lambda_0<\lambda_1<\lambda_2$. Thus, the eigenvalues and eigenvectors can be presented by
\begin{equation}
\boldsymbol{V}=\begin{bmatrix}
1 & v_{12} & v_{13} \\ 
1 & v_{22} & v_{23} \\ 
1 & v_{32} & v_{33} \\ 
\end{bmatrix},\boldsymbol{\Lambda}= \
\begin{bmatrix}
\lambda_0 & 0 & 0 \\ 
0 & \lambda_1 & 0 \\ 
0 & 0 & \lambda_2 \\ 
\end{bmatrix},
\end{equation}
with $\lambda_0=0$. Note that the corresponding eigenvector of $\lambda_0=0$ is always a vector of identical values, here normalized as $1$s.
To construct the gradient matrix of the models, we stack $\boldsymbol{g}_i$ horizontally as
\begin{equation}
\boldsymbol{G}=\begin{bmatrix}
g_{11} & g_{12} \\ 
g_{21} & g_{22} \\ 
g_{31} & g_{32}  \\  
\end{bmatrix}.
\end{equation}
The reason for stacking the gradients in the row, and not column-wise, is due to the graph's specification. For example, since $\boldsymbol{V}^T$ is a $3 \times 3$ matrix, the multiplication of $\boldsymbol{G_f}=\boldsymbol{V}^T\boldsymbol{G}$ demands the row of $\boldsymbol{G}$ to be $3$. We will further see that stacking row-wise will result in aggregating each column of $\boldsymbol{G}$. That is exactly what is required; aggregating each gradient of the model with respect to the same gradient element of another. Moreover, we set $\textrm{diag}[\kappa_1,...,\kappa_K]=\textrm{diag}[\frac{1}{N},...,\frac{1}{N}]$ to further simplify the outcome. Note that $\kappa_i$, $i=1,...,K$, privileges each device individually without any consideration of its connectivity on the graph.

Defining the filter operator as $h_s(\lambda)=\frac{1}{1+\mu_s\lambda}$, we can then derive the matrix frequency coefficients of $\boldsymbol{G}$ as  
\begin{align}
\label{eq:g_f}
\boldsymbol{G_f}&=\textrm{diag}[\kappa_1,...,\kappa_K] h_s(\boldsymbol{\Lambda}) \boldsymbol{V}^T \boldsymbol{G}=\\\nonumber
&\begin{bmatrix}
h_s(\lambda_0)\frac{1}{N} \sum_{j=1}^{3}g_{j1} & h_s(\lambda_0)\frac{1}{N} \sum_{j=1}^{3}g_{j2} \\ 
h_s(\lambda_1)\frac{1}{N} \sum_{j=1}^{3}v_{j2}g_{j1} & h_s(\lambda_1)\frac{1}{N} \sum_{j=1}^{3}v_{j2}g_{j2} \\ 
h_s(\lambda_2)\frac{1}{N} \sum_{j=1}^{3}v_{j3}g_{j1} & h_s(\lambda_2)\frac{1}{N} \sum_{j=1}^{3}v_{j3}g_{j2}  \\  
\end{bmatrix}.
\end{align}
 Each element of the columns in $\boldsymbol{G_f}$ represents a frequency coefficient. Now, let us assume we aim to construct the FedAvg algorithm using G-Fedfilt. To do so, we set $\mu_s$ to a large value. The result is $h_s(\lambda_0 = 0)=1$ and $h_s(\lambda_1)\approx h_s(\lambda_2)\approx 0$. In this case, only the first row of $\boldsymbol{G}_f$ in \eqref{eq:g_f} becomes non-zero, i.e., only DC coefficients remain. Performing the IGFT on $\boldsymbol{G}_f$ gives the aggregated gradient matrix as
\begin{equation}
\boldsymbol{\widehat{G}}=\boldsymbol{V}\boldsymbol{G}_f=\begin{bmatrix}
\frac{1}{N}\sum_{j=1}^{3}g_{j1} & \frac{1}{N}\sum_{j=1}^{3}g_{j2} \\ 
\frac{1}{N}\sum_{j=1}^{3}g_{j1} & \frac{1}{N}\sum_{j=1}^{3}g_{j2} \\ 
\frac{1}{N}\sum_{j=1}^{3}g_{j1} & \frac{1}{N}\sum_{j=1}^{3}g_{j2}  
\end{bmatrix}.
\end{equation}
It is seen that each row of $\boldsymbol{\widehat{G}}$ is an averaged version of the original gradients, similar to the behavior of the FedAvg algorithm. Note that the $i^{th}$ row belongs to the gradient update vector of the $i^{th}$ node.

\bibliographystyle{unsrtnat}
\bibliography{refs}  






\end{document}